%% file: main.tex
\documentclass[letterpaper, paper,11pt]{AAS}		

\usepackage{bm}
\usepackage{amsmath}
\usepackage{amssymb}
\usepackage[colorlinks=true, pdfstartview=FitV, linkcolor=black, citecolor= black, urlcolor= black]{hyperref}
\usepackage{overcite}
\usepackage{footnpag}			      	
\usepackage{color}
\usepackage{soul}
\usepackage{caption}
\usepackage{subcaption}
\usepackage{siunitx}
\usepackage{textcomp}
\usepackage{mathabx}
\usepackage{multirow}
\usepackage{changepage}
\usepackage{threeparttable, tablefootnote}

\def\gncnet{G\&CNet}

 \newcommand{\hlc}[2][yellow]{ {\sethlcolor{#1} \hl{#2}} }

\PaperNumber{20-616}

\begin{document}

\title{REINFORCEMENT LEARNING FOR LOW-THRUST TRAJECTORY DESIGN OF INTERPLANETARY MISSIONS}

\author{Alessandro Zavoli\thanks{Researcher, Department of Mechanical and Aerospace Engineering, Sapienza University of Rome, via Eudossiana 18, 00184, Rome, Italy}  
\ and Lorenzo Federici\thanks{PhD student, Department of Mechanical and Aerospace Engineering, Sapienza University of Rome, via Eudossiana 18, 00184, Rome, Italy}
}

\maketitle{}

\begin{abstract}
%
This paper investigates the use of Reinforcement Learning for the robust design of low-thrust interplanetary trajectories in presence of severe disturbances, modeled alternatively as Gaussian additive process noise, observation noise, control actuation errors on thrust magnitude and direction, and possibly multiple missed thrust events.
The optimal control problem is recast as a time-discrete Markov Decision Process
to comply with the standard formulation of reinforcement learning.
An open-source implementation of the state-of-the-art algorithm Proximal Policy Optimization is adopted to carry out the training process of a deep neural network, used to map the spacecraft (observed) states to the optimal control policy. The resulting Guidance and Control Network provides both a robust nominal trajectory and the associated closed-loop guidance law.
Numerical results are presented for a typical Earth-Mars mission.
First, in order to validate the proposed approach, the solution found in a (deterministic) unperturbed scenario is compared with the optimal one provided by an indirect technique.
Then, the robustness and optimality of the obtained closed-loop guidance laws is assessed by means of Monte Carlo campaigns performed in the considered uncertain scenarios. 
These preliminary results open up new horizons for the use of reinforcement learning in the robust design of interplanetary missions.
%
%

\end{abstract}

\section{Introduction}

In recent years, the possibility of using small or micro-spacecraft in interplanetary missions is drawing the attention of scientists and engineers around the world interested in reducing both development time and cost of the mission, without affecting significantly its scientific return.
The first deep-space micro-spacecraft, PROCYON \cite{campagnola2015low}, was developed in little more than a year in 2014 by the University of Tokyo and JAXA, at a very low cost if compared to standard-size spacecraft. Despite the malfunctioning of the main thruster, the PROCYON mission has been ubiquitously called a success, paving the way for similar mission concepts by other space agencies. 
In 2018, NASA released the first two interplanetary CubeSats, part of the MarCO (Mars Cube One) mission \cite{asmar2014mars}, which successfully accomplished their goal of providing a real-time communication link to Earth during the entry, descent, and landing phase of InSight lander. The same year, ESA's first stand-alone CubeSat mission for deep-space, M–Argo (Miniaturised – Asteroid Remote Geophysical Observer) has been announced \cite{walker2017miniaturised}, and it is likely to be ready for launch in mid-2021 at the earliest.

Low-thrust electric propulsion is a key technology for enabling small/micro-satellite interplanetary missions, as it provides the spacecraft with significantly lower specific propellant consumption. However, because of the limited budget, micro-spacecraft generally mount components with a low technological readiness level. This increases the risk of incurring unexpected control execution errors and/or missed thrust events (MTEs) during any of the long thrusting periods. In addition, small spacecraft have limited ground station access, and larger uncertainties in the state knowledge (i.e., in the observations for orbit determination) should be expected with respect to standard missions.

Typically, when designing the mission, the engineers take these uncertainties into account \textit{a posteriori}\cite{rayman2007coupling, laipert2015automated}, by means of time-consuming iterative procedures which often bring to suboptimal solutions and over-conservative margins. This design methodology is particularly unsuitable for micro-spacecraft missions, where the possibility to have large propellant margins and system redundancy is almost completely excluded. 
In this respect, recent works attempted to address the robust design of interplanetary trajectories by using novel optimization techniques.
As an example, the problem of designing optimal risk-aware trajectories, which guarantee the safety of the spacecraft when it operates in uncertain environments, was addressed by applying chance-constrained optimal control~\cite{ono2013probabilistic}, combined with a convex optimization approach, to deal with impulsive maneuvers~\cite{oguri2019risk}, or with a direct/indirect hybrid optimization method, to deal with continuous-thrust~\cite{oguri2019risk2}.
Stochastic Differential Dynamic Programming (SDDP) was applied to interplanetary trajectory design in presence of Gaussian-modeled state uncertainties \cite{ozaki2018stochastic, ozaki2020tube}. 
Also, the robust design of a low-thrust interplanetary transfer to a near-Earth asteroid was performed by using evidence theory to model epistemic uncertainties in the performance of the main thruster and in the magnitude of the departure hyperbolic excess velocity~\cite{dicarlo2019robust}.
Belief-based transcription procedures for the stochastic optimal control problem were proposed for the robust design of space trajectories under stochastic and epistemic uncertainties~\cite{greco2018intrusive, greco2020direct}, incorporating also navigation analysis in the formulation to update the knowledge of the spacecraft state in presence of observations~\cite{greco2020robust}.

\subsection{Deep Learning in Spaceflight Mechanics}
The interest in the application of deep learning techniques to optimally and robustly solve control problems is rapidly increasing in recent years, especially for space applications.
In this context, the term \gncnet{} (namely, Guidance and Control Network) was coined at the European Space Agency \cite{izzo2019survey} to refer to an on-board system that provides real-time guidance and control functionalities to the spacecraft by means of a Deep Neural Network (DNN) that replaces traditional control and guidance architectures.
%
{DNNs are among the most versatile and powerful machine learning tools, thanks to their unique capability of accurately approximating complex, nonlinear input-output functions, provided that a sufficiently large amount of data (training set) consisting of sample input-output pairs is available} \cite{hornik1990universal}. 
Two alternative, and quite different, approaches can be used for training a \gncnet{} to solve an optimal control problem (OCP), depending on what training data are used and how they are collected.

In \textit{Behavioral Cloning} (BC), given a set of trajectories from an expert (that is, labeled observations-controls pairs), the network is trained to reproduce (or clone) the expert behavior. 
Usually, these trajectories are obtained as the solution 
of a {(deterministic)} optimal control problem with randomized boundary conditions.
Behavioral cloning has been successfully used to train a fast-execution
\gncnet{} to control a spacecraft during a fuel-optimal low-thrust Earth-Venus transfer~\cite{izzo2019interplanetary} as well as during a landing maneuver in a simplified dynamical model~\cite{sanchez2016learning}.
This approach proved to be computationally efficient, and it benefits from state-of-the-art implementations of supervised learning algorithms \cite{tensorflow2015-whitepaper}.
However, it shows a number of downsides that make it unsuitable for robust trajectory design. 
In fact, the BC effectiveness rapidly worsens when the \gncnet{} is asked to solve problems that fall outside of the set of expert demonstrations it was trained in. As a consequence, when dealing with Stochastic Optimal Control Problems (SOCPs), a drop in performance (or even divergence) may occur when, because of uncertainty, the flight trajectory starts moving away from the training set domain, typically populated by solutions coming from deterministic OCPs.
To recover a correct behavior, 
a DAGGER (Dataset Aggregation) algorithm can be used.
In this case, the solution process features and additional loop
where new training data are provided ``on-line'' by an expert (e.g., an OCP solver)
as they are required to cover previously unknown situations.
This approach has been effectively exploited to improve the network accuracy in controlling a lander during a powered descent on the Lunar surface~\cite{furfaro2018deep}.
However, the effectiveness of BC for robust trajectory design remains doubtful, especially when solutions from deterministic OCPs are used as expert demonstrations. Recently, an attempt has been performed to train a network by BC with a training set encompassing trajectories perturbed by random MTEs \cite{rubinsztejn2019neural}, showing promising results. However, the possibility of having other types of state and control uncertainties has not been addressed yet.


A different approach is represented by \textit{Reinforcement Learning} (RL), which involves learning from experience rather than from expert demonstrations. 
In RL, a software agent (e.g., the \gncnet{}) 
autonomously learns how to behave in a (possibly) unknown dynamical environment, modeled as a Markov Decision Process (MDP),
so as to maximize some utility-based function that plays the role of the merit index in traditional optimal control problems.
Differently from the BC approach, there is no pre-assigned data set of observations-controls pairs to learn from, so the agent is not told in advance what actions to take in a given set of states. Instead, the agent is left free to explore the environment, by repeatedly interacting with a sufficiently large number of realizations of it. The only feedback the agent receives back is a numerical reward collected at each time step, which helps the agent understanding how good or how bad its current performance is.
In this framework, the final goal of the RL-agent is to learn the control policy that maximizes the expected cumulative sum of rewards over a trajectory.
Because MDP allows only scalar reward functions, a careful choice, or shaping, of the reward is mandatory to efficiently guide the agent during training, while ensuring compliance with (any) problem constraints.
Deep RL methods have obtained promising results in a number of spaceflight dynamics problems, such as low-thrust interplanetary trajectory design \cite{miller2019low, miller2019interplanetary, sullivan2020using}, 3-DoF and 6-DoF landing guidance with application to a powered descent~\cite{gaudet2020deep}, trajectory optimization in the cislunar environment \cite{scorsoglio2019actor, lafarge2020guidance}, and the design of guidance algorithms for rendezvous and docking maneuvers~\cite{broida2019spacecraft, hovell2020deep}.

This paper aims at investigating the use of Reinforcement Learning for the robust design of a low-thrust interplanetary trajectory in presence of uncertainty. Specifically, uncertainties on the spacecraft state, caused by unmodeled dynamical effects, on orbit determination, because of inaccurate knowledge, and on the applied control, due to execution errors and missed thrust events, will be considered in the present analysis.
RL has been selected as optimization algorithm since it has the clear advantage of not requiring the \textit{a priori} generation of any optimal trajectory to populate the training set,
as data are gathered by running directly the current best found control policy 
on the stochastic environment.
In this way, the agent is able to progressively improve, in an autonomous way, the performance and robustness of its control policy, in order to achieve the mission goals regardless of the uncertainties that may arise. 
This feature makes RL the ideal candidate to solve the problem at hand.
At present, most of the research encompassing RL for spacecraft trajectory design deals exclusively with deterministic environments. 
Thus, one of the main contributions of this paper is the investigation of the possible extension of RL applicability also to stochastic scenarios.

The paper is organized as follows.
First,
the optimization problem is formulated as a Markov Decision Process, and the mathematical models used to describe the state, observation, and control uncertainties acting on the system are defined. The expression of the reward function, which includes both the merit index and the problem constraints (e.g., fixed final spacecraft position and velocity), is given as well.
Next, after a brief introduction of the basic concepts and notation of Reinforcement Learning, the RL algorithm used in this work, named Proximal Policy Optimization, is described in detail. Furthermore, the configuration selected for the DNN and the values used for the algorithm hyper-parameters are reported. 
Then, numerical results are presented for the case study of the paper, that is, a time-fixed low-thrust Earth-Mars rendezvous mission. Specifically, the effect of each source of uncertainties on the system dynamics is analysed independently and the obtained results are compared in terms of trajectory robustness and optimality. Eventually, the reliability of the obtained solutions is assessed by means of Monte Carlo simulations.
A section of conclusions ends the paper.

\section{Problem Statement}


This paper investigates the use of RL algorithms for the design of robust low-thrust interplanetary trajectories.
For the sake of comparison with other research papers~\cite{ozaki2018stochastic, ozaki2020tube}, 
a three-dimensional time-fixed minimum-fuel Earth-Mars rendezvous mission is considered as a test case.
The spacecraft leaves the Earth with zero excess of hyperbolic velocity, and
it is assumed to move in a Keplerian dynamical model under the sole influence of the Sun. The mission goal is to match Mars position and velocity at final time, with minimum propellant consumption.
The values of the initial position $\bm{r}_\Earth$ and velocity $\bm{v}_\Earth$ of the Earth, the final position $\bm{r}_\Mars$ and velocity $\bm{v}_\Mars$ of Mars, the total transfer time $t_f$, the initial spacecraft mass $m_0$, and the spacecraft engine parameters (maximum thrust $T_{max}$ and effective exhaust velocity $u_{eq}$) are the same as in the paper by Lantoine and Russell \cite{lantoine2012hybrid}, and are reported in Table~\ref{tab:data}.
In all simulations, the physical quantities have been made non-dimensional by using 
as reference values the Earth-Sun mean distance
$\bar{r} = \SI{149.6e6}{km}$, the corresponding circular velocity 
$\bar{v} = \sqrt{{\mu_\Sun}/{\bar{r}}}$, and the initial spacecraft mass $\bar{m} = m_0$.

\begin{table}[htbp]
    \caption{Problem data.}
   \label{tab:data}
        \centering 
   \begin{tabular}{c c} 
      \hline 
      Variable  & Value\\
      \hline 
      $N$       & 40 \\
      $t_f,\, \si{days}$     &  $358.79$ \\
      $T_{max},\, \si{\newton}$ &  $0.50$ \\
      $u_{eq},\, \si{\kilo\meter/\second}$  &  $19.6133$ \\
      $m_0,\, \si{\kilo\gram}$ & $1000$ \\
      $\mu_\Sun,\,  \si{\kilo\meter^3/\second^2}$ & $132712440018$ \\
      $\bm{r}_\Earth,\, \si{\kilo\meter}$ & $[-140699693,\, -51614428,\, 980]^T$\\
      $\bm{v}_\Earth,\, \si{\kilo\meter/\second}$ & $[9.774596,\, -28.07828,\, 4.337725 \times 10^{-4}]^T$\\
      $\bm{r}_\Mars,\, \si{\kilo\meter}$ & $[-172682023,\, 176959469,\, 7948912]^T$\\
      $\bm{v}_\Mars,\,\si{\kilo\meter/\second}$ & $[-16.427384,\, -14.860506,\, 9.21486 \times 10^{-2}]^T$\\
      \hline
   \end{tabular}
\end{table}

The stochastic effects here considered are 
\textit{state uncertainties}, which refer to the presence of unmodeled dynamics,
\textit{observation uncertainties},
related to measurement noise and/or inaccuracies in orbital determination that lead to imperfect knowledge of the spacecraft state, and
\textit{control uncertainties}, 
which account for both random actuation errors (i.e., in the direction and magnitude of the thrust), and \textit{single} or \textit{multiple MTEs}, which correspond to null thrust occurrences.




\subsection{Markov Decision Process}
Let us briefly introduce the mathematical formulation of a generic Markov Decision Process (MDP), which is required to properly setup the mathematical framework of deep RL algorithms.
Let $\bm{s}_k \in S \subset \mathbb{R}^n$ be a vector that completely identifies the \textit{state} of the system (e.g., the  spacecraft) at time $t_k$.
In general, the complete system state at time $t_k$ is not available to the controller, 
which instead relies on an \textit{observation} vector $\bm{o}_k \in O \subset \mathbb{R}^m$. Observations might be 
affected by noise or uncertainty, and are thus written as a function of a random vector $\bm{\omega}_{o,k} \in \Omega_o \subset \mathbb{R}^{m_w}$.
The commanded \textit{action} $\bm{a}_k$ at time $t_k$ is the output of a state-feedback control policy $\pi : O \xrightarrow{} A$, that is: $\bm{a}_k = \pi(\bm{o}_k) \in A \subset \mathbb{R}^l$.
The actual \textit{control} $\bm{u}_k \in A$ differs from the commanded action due to possible execution errors, modeled as a function of a stochastic control disturbance vector $\bm{\omega}_{a,k} \in \Omega_a \subset \mathbb{R}^{l_w}$.
A stochastic, time-discrete dynamical model $f$ is considered for the system state. The uncertainty on the system dynamics at time $t_k$ is modeled as a random vector $\bm{w}_{s,k} \in \Omega_s \subset \mathbb{R}^{n_w}$.
%
%
As a result, the dynamical system evolution over time is described by the following equations:
\begin{align}
        \bm{s}_{k+1} &= f(\bm{s}_k, \bm{u}_k, \bm{\omega}_{s, k})   \label{eq:MDP} \\
        \bm{o}_{k} &= h(\bm{s}_{k}, t_k, \bm{\omega}_{o, k}) \label{eq:MDP4} \\
        \bm{u}_k &= g(\bm{a}_k, \bm{\omega}_{a,k}) \label{eq:MDP3}\\
        \bm{a}_k &= \pi(\bm{o}_k)   \label{eq:MDP2} 
\end{align}

The problem goal is to find the optimal control policy $\pi^\ast$ that maximizes the expected value of the discounted sum of rewards, that, in an episodic form, is:
\begin{equation}
    J = \underset{\tau\sim \pi}{\mathbb{E}} \left[ \sum_{k = 0}^{N-1} { \gamma^k R(\bm{s}_k, \bm{u}_k, \bm{s}_{k+1}) }  \right]
    \label{eq:obj}
\end{equation}
where $R(\bm{s}_k, \bm{u}_k, \bm{s}_{k+1})$ is the reward associated with transitioning from state $\bm{s}_k$ to state $\bm{s}_{k+1}$ due to control $\bm{u}_k$,
$\gamma \in (0,1]$ is a discount factor that is used to either encourage long term planning ($\gamma = 1$) or short term  rewards ($\gamma \ll 1$), and $N$ is the number of steps in one episode.
Note that  
$\mathbb{E}_{\tau\sim \pi}$ here
denotes the expectation taken over a trajectory $\tau$, that is, a sequence of state-action pairs $\tau = \left\{ (\bm{s}_0,\,\bm{a}_0) ,\, \ldots \, (\bm{s}_{N-1},\,\bm{a}_{N-1}) \right\}$ 
sampled according to the closed-loop dynamics in Eqs.~\eqref{eq:MDP}-\eqref{eq:MDP2}.

Note that, in an episodic setting, $J = V^\pi(\bm{s}_0)$, being $V^\pi(\bm{s}_k)$ the value function, defined as the expected return obtained by starting from state $\bm{s}_k$ and acting according to policy $\pi$ until the end of the episode:
\begin{equation}
    V^{\pi}(\bm{s}_k) = \underset{\tau\sim \pi}{\mathbb{E}} \left[ \sum_{k' = k}^{N-1} { \gamma^{k'} R(\bm{s}_{k'}, \bm{u}_{k'}, \bm{s}_{k'+1}) }  \right]
    \label{eq:Vpi}
\end{equation}

\subsection{Formulating an Earth-Mars Mission as a Markov Decision Process}

This general model is now specified for the Earth-Mars transfer problem at hand.
During the mission, the spacecraft state $\bm{s}_k$ at any time step $t_k = k\, t_f / N,\, k \in [0,N]$, is identified by its inertial position $\bm{r}$ and velocity $\bm{v}$ with respect to Sun, and by its total mass $m$: 
\begin{equation}
    \bm{s}_k = \left[\bm{r}_k^T, \bm{v}_k^T, m_k \right]^T \in \mathbb{R}^7
\end{equation}
The low-thrust trajectory is approximated as a series of ballistic arcs connected by impulsive $\Delta V$s, similarly to what done in the well-known Sims-Flanagan model~\cite{sims1999preliminary}. The magnitude of the $k$-th impulse is limited by the amount of $\Delta V$ that could be accumulated over the corresponding trajectory segment by operating the spacecraft engine at maximum thrust $T_{max}$:
\begin{equation}
    \Delta V_{max, k} = \frac{T_{max}}{m_k} \frac{t_f}{N}
    \label{eq:DVmax-k}
\end{equation}
So, the commanded action at time $t_k$ corresponds to an impulsive $\Delta V$: 
\begin{equation}
    \bm{a}_k = \Delta \bm{V}_k \in [-\Delta V_{max, k}, \Delta V_{max, k}]^3 \subset \mathbb{R}^3.
    \label{eq:action}
\end{equation}
Since the spacecraft moves under Keplerian dynamics between any two time steps, in a deterministic scenario the spacecraft state can be propagated analytically with a closed-form transition function:
\begin{equation}
    \begin{bmatrix}
    \bm{r}_{k+1} \\
    \bm{v}_{k+1} \\
    m_{k+1} \\
    \end{bmatrix}
    = f(\bm{r}_k, \bm{v}_k, m_k, \Delta \bm{V}_k) = 
    \begin{bmatrix}
    \hat f_k \bm{r}_k + \hat g_k (\bm{v}_k + \Delta \bm{V}_k) \\
    \dot{\hat f}_k \bm{r}_k +  \dot{\hat g}_k (\bm{v}_k + \Delta \bm{V}_k)\\
    m_k \, \mbox{exp}\left({-\frac{|\Delta \bm{V}_k|}{u_{eq}}}\right) \\
    \end{bmatrix}
\label{eq:dyn_kep}
\end{equation}
where $\hat f_k$ and $\hat g_k$ are the Lagrange coefficients at $k$-th step, defined as in Ref.~\citenum{bate1971fundamentals}, and the mass update is obtained through Tsiolkovsky equation.

At time $t_f$, the final $\Delta V$ is calculated so as to match Mars velocity, that is: \begin{equation}
    \Delta \bm{V}_N = \min{\left(|\bm{v}_\Mars - \bm{v}_N|, \Delta V_{max,N}\right)} \frac{\bm{v}_\Mars - \bm{v}_N}{|\bm{v}_\Mars - \bm{v}_N|}
    \label{eq:DVmax-f}
\end{equation}
and the final spacecraft state is evaluated as:
\begin{align}
    \bm{r}_f &= \bm{r}_N \\
    \bm{v}_f &= \bm{v}_N + \Delta \bm{V}_N \\
    m_f &=  m_N \, \mbox{exp}\left({-{|\Delta \bm{V}_N|}/{u_{eq}}}\right)
\end{align}

The (deterministic) observations collected at time $t_k$ are:
\begin{equation}
    \bm{o}_k = \left[\bm{r}_k^T, \bm{v}_k^T, m_k, t_k \right]^T \in \mathbb{R}^8
\end{equation}

The value selected for the total number of time steps $N$ is reported in Table~\ref{tab:data}.

\subsubsection{State Uncertainties.}
For the sake of simplicity, uncertainties on the spacecraft dynamics are modeled as additive Gaussian noise on position and velocity at time $t_k$, $k \in (0, N]$, that is:
\begin{equation}
    \bm{w}_{s, k} = 
    \begin{bmatrix}
    \delta \bm{r}_k \\
    \delta \bm{v}_k
    \end{bmatrix}
    \sim \mathcal{N}(\bm{0}_{6}, \bm{R}_{s,k}) \in \mathbb{R}^6
\end{equation}
where $\bm{R}_{s,k} = \mbox{diag}\left(\sigma_r^2\bm{I}_3, \sigma_v^2\bm{I}_3 \right)$ is the covariance matrix,
$\bm{I}_{n}$ (respectively, $\bm{0}_{n}$) indicates an identity (respectively, null) matrix with dimension $n \times n$ (respectively, $n \times 1$), and $\sigma_r, \sigma_v$ are the standard deviations on position and velocity.
So, the stochastic dynamical model is written as:
\begin{equation}
    \begin{bmatrix}
    \bm{r}_{k+1} \\
    \bm{v}_{k+1} \\
    m_{k+1} \\
    \end{bmatrix}
    = f(\bm{r}_k, \bm{v}_k, m_k, \bm{u}_k) + 
    \begin{bmatrix}
    \delta \bm{r}_{k+1} \\
    \delta \bm{v}_{k+1} \\
    0 \\
    \end{bmatrix}
\end{equation}

\subsubsection{Observation Uncertainties.}
The uncertainty in the knowledge of spacecraft position and velocity due to errors in the orbital determination is modeled as additive Gaussian noise on the deterministic observations at time $t_k$:
\begin{equation}
    \bm{o}_k = 
    \begin{bmatrix}
    \bm{r}_{k} \\
    \bm{v}_{k} \\
    m_{k} \\
    t_k
    \end{bmatrix}
    +
    \begin{bmatrix}
    \delta \bm{r}_{o, k} \\
    \delta \bm{v}_{o, k} \\
    0 \\
    0
    \end{bmatrix}
\end{equation}
being:
\begin{equation}
    \bm{w}_{o, k} = 
    \begin{bmatrix}
    \delta \bm{r}_{o,k} \\
    \delta \bm{v}_{o,k}
    \end{bmatrix}
    \sim \mathcal{N}(\bm{0}_{6}, \bm{R}_{s,k}) \in \mathbb{R}^6
\end{equation}

\subsubsection{Control Uncertainties.}
Control execution errors are modeled as a small three-dimensional rotation of the commanded $\Delta V$ vector, defined by Euler angles $(\delta \phi, \delta \vartheta, \delta \psi)$, and a slight variation $\delta u$ of its modulus. Random variables $\delta \phi, \delta \vartheta, \delta \psi$ and $\delta u$ are assumed to be Gaussian, with standard deviations $\sigma_{\phi}, \sigma_{\vartheta}, \sigma_{\psi}$ and $\sigma_{u}$. So, the control disturbance vector at time $t_k$ is:
\begin{equation}
    \bm{w}_{a, k} = 
    \begin{bmatrix}
    \delta \phi_k \\
    \delta \vartheta_k \\
    \delta \psi_k \\
    \delta u_k
    \end{bmatrix}
    \sim \mathcal{N}(\bm{0}_{4}, \bm{R}_{a,k}) \in \mathbb{R}^4
\end{equation}
where $\bm{R}_{a,k} = \mbox{diag}\left(\sigma_\phi^2, \sigma_\vartheta^2, \sigma_\psi^2, \sigma_u^2 \right)$ is the covariance matrix.

The actual control $\bm{u}$ can be written as a function of the commanded action $\bm{a}$ at time $t_k$, $k \in [0, N)$, as:
\begin{equation}
    \bm{u}_k = g(\bm{a}_k, \bm{w}_{a,k}) = (1 + \delta u_k) \bm{A}_k \bm{a}_k
\end{equation}
where the rotation matrix $\bm{A}_k$ is evaluated, under the  small-angle assumption, as:
\begin{equation}
    \bm{A}_k = 
    \begin{bmatrix}
    1 & - \delta \psi_k & \delta \vartheta_k\\
    \delta \psi_k & 1 & - \delta \phi_k \\
    - \delta \vartheta_k & \delta \phi_k & 1
    \end{bmatrix}
\end{equation}
It is worth noting that, although the control disturbance vector is Gaussian, the effect obtained on the applied control is definitively non-Gaussian and, for this reason, the solution methods in Ref.~\citenum{ozaki2018stochastic} and \citenum{ozaki2020tube} may not be applicable. 

\subsubsection{Missed Thrust Events.}
Besides small control execution errors, the effect of  one or more consecutive MTEs over the course of the mission is also investigated.
The MTE is modeled as a complete lack of thrust, even when commanded, that occurs at a randomly chosen time  $t_{\hat k} \in [0, N)$, so that $\bm{u}_{\hat k}=\bm{0}_{3}$.
With some probability $1-p_{mte}$ the miss-thrust is recovered and for the remaining steps it never happens again.
Otherwise, the MTE persists for an additional time-step. This procedure is repeated, but the MTE may last at most $n_{mte}$ successive time steps, that is, from $t_{\hat k}$ to $t_{\hat k +n_{mte}-1}$.
%
%

The values used for the standard deviations and the other uncertainty model parameters introduced so far are reported in Table~\ref{tab:sigma}.

\begin{table}[htbp]
    \caption{Uncertainty model parameters.}
   \label{tab:sigma}
        \centering 
   \begin{tabular}{c c c c c c c c} 
      \hline
      $\sigma_r,\, \si{\kilo\meter}$       & 
      $\sigma_v,\, \si{\kilo\meter/\second}$     &   
      $\sigma_\phi,\, \si{deg}$ &
      $\sigma_\vartheta,\, \si{deg}$  & 
      $\sigma_\psi,\, \si{deg}$ & 
      $\sigma_u$ & 
      $p_{mte}$ &
      $n_{mte}$ \\
      \hline
      $1.0$  & $0.05$ &  $1.0$ &  $1.0$ & $1.0$  & $0.05$ & $0.1$ & $3$ \\
      \hline 
   \end{tabular}
\end{table}


\subsubsection{Reward Function.}
The objective of the optimization procedure is to maximize the (expected) final mass of the spacecraft, while ensuring the compliance with terminal rendezvous constraints on position and velocity.
For this reason, the reward $r_k$ collected by the agent at time $t_k$, for $k \in (0, N]$, is defined as:
\begin{equation}
    r_k = -\mu_k -\lambda_1 \, e_{u,k-1} - \lambda_2 \, e_{s,k}
    \label{eq:rew}
\end{equation}
where:
\begin{align}
\mu_k &= \Delta m_k = 
     \begin{cases}
    m_{k-1} - m_k & \mbox{if}\,\, k < N \\
    m_{N-1} - m_f & \mbox{if}\,\, k = N
    \end{cases} \\
    e_{u,k} &= \max{\left(0, |\bm{u}_{k}| - \Delta V_{max,k} \right)} \\
    e_{s,k} &= 
    \begin{cases}
    0 & \mbox{if}\,\, k < N \\
    \max{\left(0, \max{\left(\frac{|\bm{r}_f - \bm{r}_\Mars|}{|\bm{r}_\Mars|}, \frac{|\bm{v}_f - \bm{v}_\Mars|}{|\bm{v}_\Mars|}\right)  - \varepsilon } \right)}   & \mbox{if}\,\, k = N 
    \end{cases}  
\end{align}
%
Here $\mu_k$  is the cost function, that is, the consumed propellant mass, %
$e_{u,k}$ is the violation of the constraint relative to the maximum $\Delta V$ magnitude admissible for that segment (see Eq.~\ref{eq:DVmax-k}), and
$e_{s,k}$ is the violation of the constraint acting on the final state of the spacecraft, up to a given tolerance $\varepsilon = \SI{e-3}{}$.
The  penalty factors $\lambda_1 = 100$ and $\lambda_2=50$ are used in the present work.


\section{Reinforcement Learning}
\input{RL}


\section{Numerical Results}
\input{results.tex}

\section{Conclusion}
This paper presented a deep Reinforcement Learning (RL) framework to deal with the robust design of low-thrust interplanetary trajectories in presence of different sources of uncertainty.
The stochastic optimal control problem must first be reformulated as a Markov Decision Process.
Then, a state-of-the-art RL algorithm, named Proximal Policy Optimization (PPO), is adopted for the problem solution, and its prominent features over similar policy-gradient methods are outlined.
Preliminary numerical results were reported for a three-dimensional Earth-Mars mission, by considering separately the effect of different types of uncertainties, namely, uncertainties on the dynamical model, on the observations, on the applied control, as well as the presence of a single or multiple, consecutive, missed thrust events.

The obtained results show the capability of PPO of solving simple interplanetary transfer problems, as the Earth-Mars mission here considered, in both deterministic and stochastic scenarios. The solution found in the deterministic case is in good agreement with the optimal solution provided by an indirect method. However, the high computational cost necessary to train the neural network discourages the use of a model-free RL algorithm in that circumstance.
The power of RL becomes apparent when dealing with stochastic optimal control problems, where traditional 
methods are either cumbersome, impractical, or simply impossible to apply.
Despite the reported results are only preliminary, the presented solutions seem very promising, in terms of both payload mass and constraint enforcement.
The methodology here proposed is quite general and can be implemented, with the appropriate changes, to cope with a variety of spacecraft missions and uncertainty models. Also, extensions to arbitrary stochastic dynamical models (e.g., with possibly complex non-Gaussian perturbations) are straightforward.
This is a major advantage with respect to other techniques presented in the literature based on ad-hoc extensions of traditional optimal control methods.

The preliminary results here proposed pave the way for reinforcement learning approaches in robust design of interplanetary trajectories. 
Additional work is obviously necessary in order to increase both the efficiency of the learning process and the reliability of the solutions.
The high computational cost calls for the use of asynchronous algorithm, where the two processes of policy-rollout (for collecting experience) and policy-update (for learning) run in parallel, so as to exploit at best the massive parallelization allowed by high performance computing clusters.
Also, the use of recurrent neural networks should be investigated when dealing with non-Markov dynamical processes, as in the case of partial observability and multiple, correlated, missed thrust events. 
However, the most crucial point seems to be enhancing the constraint-handling capability of RL algorithms.
The adoption of the $\varepsilon$-constraint relaxation is a modest contribution that goes in that direction. More advanced formulations of the problem, such as Constrained Markov Decision Process (CMDP), should be investigated in the future for this purpose.



\bibliographystyle{AAS_publication}   
\bibliography{references}   
\end{document}

%% file: RL.tex
This section briefly outlines the mathematical framework of the RL method
selected to tackle the problem outlined in the previous section.
The RL algorithm adopted in this work is \textit{Proximal Policy Optimization}~\cite{schulman2017proximal}, which is a model-free policy-gradient actor-critic method, widely recognized for the high performance demonstrated on a number of continuous and high-dimensional control problems.

A model-free approach is particularly suited to design a robust trajectory. In fact, model-free RL methods do not rely on an \textit{explicit} knowledge of the MDP (Eqs.~\eqref{eq:MDP}-\eqref{eq:MDP2}) for returning the optimal control policy $\pi^\ast$.
This fact leaves us the freedom to use complex, and possibly non-analytical,
expressions for the transition function $f$, control model $g$, observation model $h$, and/or disturbance distributions, without requiring any change in the algorithm.

Policy-gradient algorithms are a common choice in RL.
The underlying idea is to directly learn the policy $\pi(\bm{s})$ that maximizes the performance index $J$. 
A stochastic policy is generally considered, as more robust and oriented toward exploration than deterministic ones.
In this sense, 
$\pi(\bm{s})$ should be intended as a shorthand for $\pi(\bm{a} | \bm{s})$, that is, the probability of choosing action $\bm{a}$, conditioned by being the system in the state $\bm{s}$.

In order to cope with complex environments and possibly large (and continuous) state and/or action spaces,  the policy $\pi$ is usually modeled by a DNN {with} parameters (i.e., weights and biases) $\theta$, and referred to as $\pi_\theta(\bm{a} | \bm{s})$ to make the dependence on the network parameters explicit.
A typical DNN is composed by a sequence of \textit{layers}, each made up of a number of basic units, called \textit{neurons}. 
A Multi-Layer Perceptron (MLP) is adopted in the present work:
each neuron receives input signals from the units in the previous layer, generates an output signal as a linear combination of the inputs, elaborated through a (typically nonlinear) \textit{activation function}, and sends this signal to the units in the next layer.
More complex architectures, such as Recurrent Neural Networks (RNNs), are also used in deep RL to cope with partially-observable or non-Markov environments, but they are not considered in this work.
%
In the case of a deterministic policy, 
the network directly outputs the action, given the current system state as input; instead, when a stochastic policy is considered, the network returns some parameters 
of the underlying probability distribution,
such as mean and variance for a Gaussian distribution, which is the most frequently adopted.
The actual action is then sampled from this probability distribution.

Once the network architecture (i.e., number of layers, layer density, and activation functions) is assigned, the problem reduces to the search of the parameters $\theta^\ast$ that maximize the merit index $J$:
\begin{equation}
\theta^\ast = \mbox{arg} \max_\theta J(\theta) = \mbox{arg} \max_\theta \left( \underset{\tau\sim \pi_\theta}{\mathbb{E}} \left[\sum_{k=0}^{N-1} r_k \right] \right)
\label{eq:grad_pol}
\end{equation}
where $r_k = R(\bm{s}_k, \bm{u}_k, \bm{s}_{k+1})$ is the immediate reward collected after taking action $\bm{a}_k \sim \pi_\theta( \cdot | \bm{s}_k)$.

In order to solve Eq.~\eqref{eq:grad_pol}, policy-gradient algorithms perform a (stochastic) gradient ascent update on $\theta$, that is: $\theta \leftarrow \theta + \alpha \nabla_\theta J(\theta)$,
where $\alpha$ is the \textit{learning rate}, which is a constant, or slowly decreasing, user-defined hyper-parameter that controls the step-size of the gradient update.

The \textit{Policy Gradient Theorem}~\cite{sutton2018reinforcement} is used to rewrite the gradient $\nabla_\theta J(\theta)$ in a more suitable form:
\begin{equation}
    \nabla_\theta  J(\theta) = \underset{\tau\sim \pi_\theta}{\mathbb{E}} \left[ \sum_{k=0}^{N-1} \nabla_\theta \log \pi_\theta(\bm{a}_k | \bm{s}_k)\,  Q^{\pi_\theta}(\bm{s}_k,\bm{a}_k) \right]
    \label{eq:nablaJ}
\end{equation}
where $Q^{\pi_\theta}(\bm{s},\bm{a}) = \mathbb{E}_{\tau \sim \pi_\theta} 
\left[\sum_{k'=k}^{N-1} r_{k'}\, |\, \bm{s}_k = \bm{s}, \bm{a}_k = \bm{a} \right]$ is the Action-Value function, or simply Q-function, that is, the expected return obtained by starting in state $\bm{s}$, taking action $\bm{a}$ and then acting according to $\pi_\theta$.



The Q-function might be estimated by a Monte-Carlo approach, using the average return over episode samples. While unbiased, this estimate has a high variance, which makes this approach unsuitable for practical purposes.
An improved solution relies on approximating the Q-value function by a second DNN, leading to the so called Actor-Critic method.
The two neural networks run in parallel and are concurrently updated:
the Actor, which returns the parametrized policy $\pi_\theta$, is updated by gradient ascent on the policy-gradient;
the Critic, which returns the parametrized Q-function $Q_\phi$,
is recursively updated using Temporal Difference~\cite{sutton2018reinforcement}.
Intuitively, the Actor controls the agent behavior while the Critic evaluates the agent performance and gives a feedback to the Actor in order to efficiently update the policy. 

To further improve the stability of the learning process and reduce variance in the sample estimate for the policy gradient, it is possible to subtract a function of the system state only, called baseline $b(\bm{s})$, from the expression~\eqref{eq:nablaJ} of the policy gradient, without changing it in expectation.
The most common choice of baseline is the value function $V^{\pi_\theta}(\bm{s})$, which, subtracted from the Action-Value function $Q^{\pi_\theta}(\bm{s},\bm{a})$, leads to the definition of the advantage function $A^{\pi_\theta}(\bm{s},\bm{a}) = Q^{\pi_\theta}(\bm{s},\bm{a}) - V^{\pi_\theta}(\bm{s})$, which expresses by how much the total reward improves by taking a specific action $\bm{a}$ in state $\bm{s}$, instead of randomly selecting the action according to $\pi_\theta(\cdot|\bm{s})$.
From a computational point of view, the advantage function is usually computed by using the so-called generalized advantage estimation:
\begin{equation}
    \hat{A}_{k} = \sum_{k'= k}^{N-1} (\gamma \lambda)^{k'-k} \delta_{k'}
    \label{eq:advantage}
\end{equation}
where: 
\begin{equation}
   \delta_k = r_k + \gamma V^{\pi_\theta}(\bm{s}_{k+1}) - V^{\pi_\theta}(\bm{s}_{k}) 
   \label{eq:TDerror}
\end{equation}
is the Temporal-Difference error, that is an unbiased estimate of the advantage function, being ${\mathbb{E}}_{\tau \sim \pi_\theta}[\delta_k] = A^{\pi_\theta}(\bm{s}_k,\bm{a}_k)$.

So, the policy gradient can be evaluated as:
\begin{equation}
    \nabla_\theta  J(\theta) = \underset{\tau\sim \pi_\theta}{\mathbb{E}} \left[ \sum_{k=0}^{N-1} \nabla_\theta \log \pi_\theta(\bm{a}_k | \bm{s}_k)\,  \hat{A}_{k} \right]
    \label{eq:nablaJ_A}
\end{equation}

\begin{figure} [htb]
    \centering
    \includegraphics[width = 0.6\textwidth]{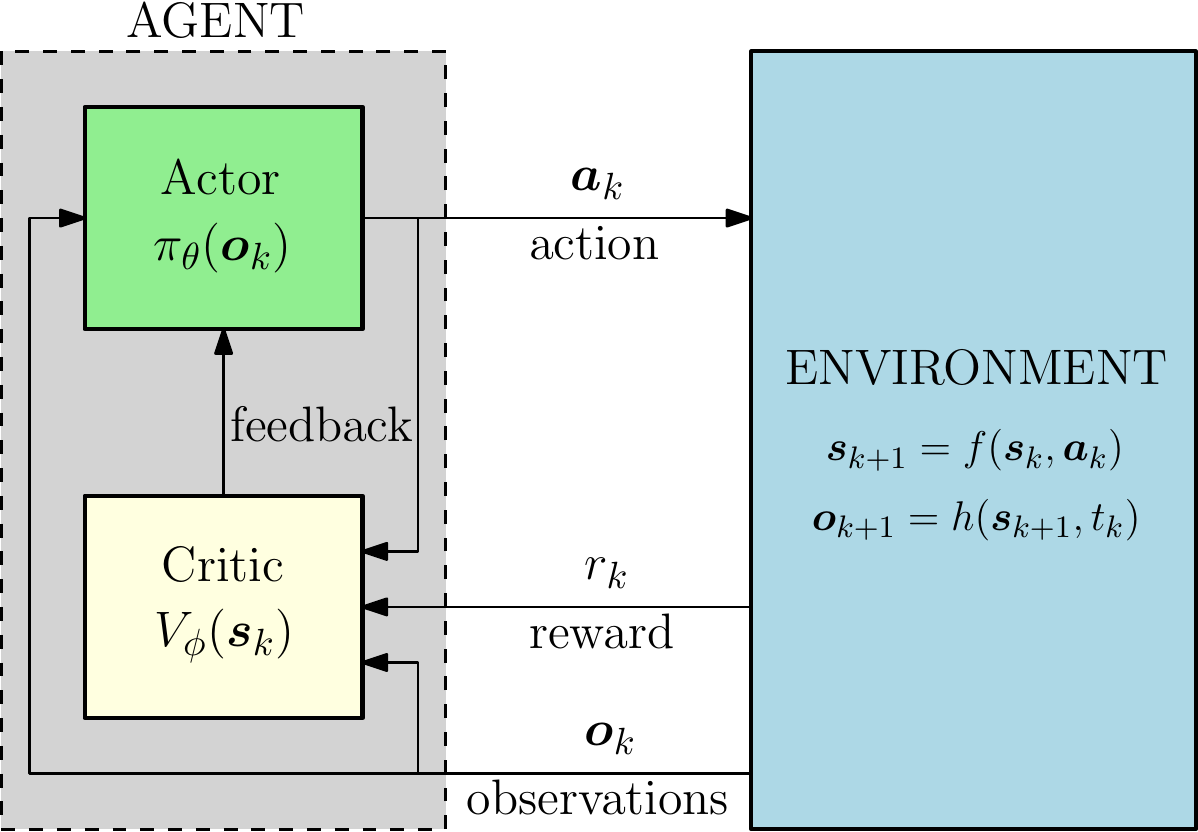}
    \caption{Schematic of the Advantage Actor-Critic RL process, for a deterministic MDP.}
    \label{fig:AC}
\end{figure}

This leads to the Advantage Actor-Critic (A2C) method \cite{mnih2016asynchronous}, whose basic architecture is reported in Fig.~\ref{fig:AC}. In this case, the Critic returns a parameterized value function $V_\phi$, necessary for the advantage estimation in Eqs.~\eqref{eq:advantage}--\eqref{eq:TDerror}.

PPO can be seen as a further evolution of the (advantage) Actor-Critic  method.
In order to avoid too large policy updates when using Eq.~\eqref{eq:nablaJ_A} during gradient ascent, which usually cause performance collapse, PPO introduces a ``clipped surrogate objective function'' $J^{clip}$ in place of $J$, to constrain the updated policy $\pi_{\theta}$ to stay in a small range $\epsilon$, named clip range, around the previous value $\pi_{\theta_{old}}$, by clipping the probability that the new policy moves outside of the interval $[1-\epsilon, 1+\epsilon]$.

Let $\tilde{r}_k(\theta)$ be the probability ratio:
\begin{equation}
    \tilde{r}_k(\theta) = \frac{\pi_{\theta}(\bm{a}_k | \bm{s}_k)}{\pi_{\theta_{old}}(\bm{a}_k | \bm{s}_k)}
\end{equation}
The clipped surrogate objective function can be written as:
\begin{equation}
    J^{clip}(\theta) =  \underset{\tau \sim \pi_{\theta}}{\mathbb{E}}\left[ \frac{1}{N} \sum_{k = 0}^{N-1}  \min{\left( \tilde{r}_k(\theta) \hat{A}_k, \mbox{clip}(\tilde{r}_k(\theta), 1-\epsilon, 1+\epsilon) \hat{A}_k\right)} \right]
\end{equation}

Empirical evidences suggest that using a clip range $\epsilon$ that decreases linearly with the training steps improves the learning effectiveness.
Typically, in PPO the policy and value function updates are carried out all at once by gathering in the set of parameters $\theta$ the weights and biases of both the Actor and Critic networks, and by including  in the objective function a mean-squared value function error term $H$:
\begin{equation}
    H(\theta) = \underset{\tau \sim \pi_{\theta}}{\mathbb{E}}\left[ \frac{1}{N} \sum_{k = 0}^{N-1} \frac{1}{2} \left(V_\theta (\bm{s}_k) - \sum_{k' = k}^{N} r_{k'} \right)^2 \right]
\end{equation}
and an entropy term $S$:
\begin{equation}
S(\theta) = \underset{\tau \sim \pi_{\theta}}{\mathbb{E}}\left[ \frac{1}{N} \sum_{k = 0}^{N-1} \mathbb{E}_{ 
\bm{a} \sim \pi_\theta(\cdot| \bm{s}_k)} \left[ - \log{ \pi_\theta(\bm{a} | \bm{s}_k)} \right]  \right]
\end{equation}
which is usually added to ensure sufficient exploration, and prevent premature convergence to a suboptimal deterministic policy \cite{mnih2016asynchronous}. 
Eventually, the final surrogate objective function is:
\begin{equation}
    J^{ppo}(\theta) = J^{clip}(\theta) - c_1 H(\theta) + c_2 S(\theta)
\end{equation}
where
$c_1$ and $c_2$ are two hyper-parameters, named value function coefficient and entropy coefficient, that control the relative importance of the various terms. 
 
The overall learning process consists of two well-distinguished phases that are repeated iteratively:
i) \textit{policy rollout}, 
during which the current policy $\pi_{\theta}$ is run
in $n_{env}$ independent (e.g. parallel) realizations of the environment for $n_b$ training episodes, collecting a set of $n_{env} n_b$  trajectories $\tau^i$; 
ii) \textit{policy update}, 
which is performed by running $n_{opt}$ epochs of stochastic gradient ascent over $n_b$ sampling mini-batches
of on-policy data, that is, the data coming from the last rollout only.
The algorithm terminates after a total number of training steps equal to $T$.


\subsubsection{Remark.} 
The above described framework has been derived in case of a (perfectly observable) Markov Decision Process.
When a perfect knowledge of the state is not available
or when the observations differ from the state, the same RL algorithm can be used, but in this case the Actor and Critic networks take as an input directly the observation $\bm{o}_k$.

\subsection{Implementation Details}%
The results presented in this work have been obtained by using the PPO algorithm implementation by Stable Baselines \cite{stable-baselines2018}, an open-source library containing a set of improved implementations of RL algorithms based on OpenAI Baselines.
The scientific library \textit{pykep}\cite{izzo2012pykep}, developed at the European Space Agency, was instead used for the astrodynamics routines.

The selected \gncnet{} consists of two separate head networks, one for the control policy and the other for the value function, each one composed of two hidden layers. The \gncnet{} architecture is summarized in Table~\ref{tab:net}, which reports the number of neurons in each layer and the activation function used in each neuron.
The tuning of PPO hyper-parameters was performed by using Optuna~\cite{optuna2019}, an open source hyper-parameter optimization framework for automated hyper-parameter search. The hyper-parameter optimization was realized on a deterministic version of the RL environment, with a budget of 500 trials, each with a maximum of $\SI{3e5}{}$ training steps. The optimal value of the hyper-parameters used in all simulations is reported in Table~\ref{tab:hyper}.




\begin{table}[!htb]
     \begin{minipage}{.45\textwidth}
        \centering
        \caption{Network Configuration.}\label{tab:net}
        \centering
        \begin{tabular}{c  c c} 
          \hline 
           & Policy  & Value \\
          & network & network\\
           \hline 
          Layer 1 & 64 & 64\\
          Layer 2 & 64 & 64\\
          Output & 3 & 1 \\
          Activation & tanh & tanh \\
          \hline 
       \end{tabular} 
    \end{minipage}
    \hfill
    \begin{minipage}{.45\textwidth}
    \begin{threeparttable}
             \caption{PPO hyperparameters.} \label{tab:hyper}
        \centering
       \begin{tabular}{c c} 
          \hline 
          Hyper-parameter  & Value\\
          \hline
          $\gamma$       & $0.9999$ \\
          $\lambda$ & $0.99$ \\
          $\alpha$     &  $2.5 \times 10^{-4} \left(1 - {t}/{T} \right)$\tnote{$\star$} \\
          $\epsilon$ &  $0.3 \left(1 - {t}/{T}\right)$\tnote{$\star$} \\
          $c_1$  &  $0.5$ \\
          $c_2$ & $4.75 \times 10^{-8}$ \\
          $n_{opt}$ & $30$\\
          \hline
       \end{tabular}
       \begin{tablenotes}\footnotesize
        \item [$\star$] $t$ indicates the training step number
        \end{tablenotes}
       \end{threeparttable}
    \end{minipage}%
%
\end{table}

When dealing with constrained optimization problems, constraint violations are typically included as penalty terms inside the reward function (see Eq.~\eqref{eq:rew}). In these cases, a penalty-based $\varepsilon$-constraint method, similar to those sometimes used in stochastic global optimization \cite{federici2020eos}, proved to be helpful to enforce constraints more gradually during optimization, allowing the agent to explore to a greater extent the solution space  at the beginning of the training process.
For this reason, as a modest original contribution of this paper, the constraint satisfaction tolerance $\varepsilon$ also varies during the training, according to a (piecewise constant) decreasing trend:
\begin{equation}
\varepsilon = 
    \begin{cases}
    0.01 & \mbox{for the first } T/2  \mbox{ training steps}\\
    0.001 & \mbox{for the second } T/2  \mbox{ training steps}
    \end{cases}
\end{equation}

%% file: results.tex
This section presents some 
preliminary results obtained by training the \gncnet{} on different environments (that is, mission scenarios), generated by considering separately each uncertainty source defined in the problem statement.
In order to validate the proposed approach, the solution found in a (deterministic) unperturbed scenario is compared with the optimal one provided by an indirect technique.
Then, the robustness and optimality of the obtained control policies is assessed by means of Monte Carlo campaigns performed in the considered uncertain scenarios.

 
\subsection{Deterministic Optimal Trajectory}
The ability of the presented methodology to deal with traditional, deterministic, optimal control problems is investigated first, by comparing the solution provided by the control policy $\pi^{unp}$ trained in the deterministic (unperturbed) environment (see Eq.~\eqref{eq:dyn_kep}), 
with the solution of the original Earth-Mars low-thrust transfer problem, found by using a well-established indirect optimization method used by the authors in other interplanetary transfers \cite{colasurdo2014tour}.
The two solutions are very close to each other in terms of trajectory and control direction.
Also, the final mass of the RL solution ($\SI{600.23}{kg}$) is in good agreement with the (true) optimal mass obtained by the indirect method ($\SI{599.98}{\kilo\gram}$).
This slight difference is partly due to the fact that RL satisfies the terminal constraints with a lower accuracy  ($10^{-3}$ in RL vs $10^{-9}$ in the indirect method), and partly due to the (approximated) time-discrete, impulsive dynamical model adopted in the MDP transcription.
%

However, when applying RL to the solution of deterministic optimal control problems, two major downsides arise.
First, terminal constraints cannot be explicitly accounted for, 
and constraint violations must be introduced in the reward function as (weighted) penalty terms. As a result,
the accuracy on constraint satisfaction is generally looser than in traditional methods for solving optimal control problems.
Second, RL is quite computationally intensive, even when applied to problems as simple as the deterministic rendezvous mission here considered. This is mainly motivated by the fact that PPO is a model-free algorithm, hence, the knowledge of the underlying (analytical) dynamical model is not exploited at all. The only way the agent can obtain satisfactory results is to acquire as much experience (i.e., samples) as possible about the environment.
In this respect, the solution of the deterministic problem here reported took about $2\div3$ hours (depending on the desired accuracy on the constraints) on a computer equipped with Intel Core i7-9700K CPU @\SI{3.60}{\giga\Hz}, while the indirect method just a few seconds.

\subsection{Robust Trajectory Design}

%
Besides the unperturbed, deterministic  mission scenario (labeled ${unp}$),
the following stochastic case-studies are considered:
i) state uncertainties ($st$), 
ii) observation uncertainties ($obs$),
iii) control uncertainties ($ctr$),
iv) single missed thrust event ($mte,1$), and 
v) multiple missed thrust events ($mte,2$).
Training the \gncnet{} in one of these environments 
leads to the definition of a corresponding policy, named
$\pi^{unp}$, $\pi^{st}$, $\pi^{obs}$, $\pi^{ctr}$, $\pi^{mte,1}$, and $\pi^{mte,2}$, respectively.
%
%
For each policy, the reference trajectory, which should be intended as robust to that source of uncertainty, is obtained by applying in the unperturbed environment a (deterministic) version of the policy (i.e., that always takes the action corresponding to the largest probability, instead of sampling from the probability distribution $\bm{a}\sim\pi(\cdot|\bm{s})$), and recording the commands and spacecraft states along the flight.

\begin{figure} [!htbp]
    \centering
    \includegraphics[width=0.75\textwidth, trim={0cm 0.5cm 0cm 1cm},clip]{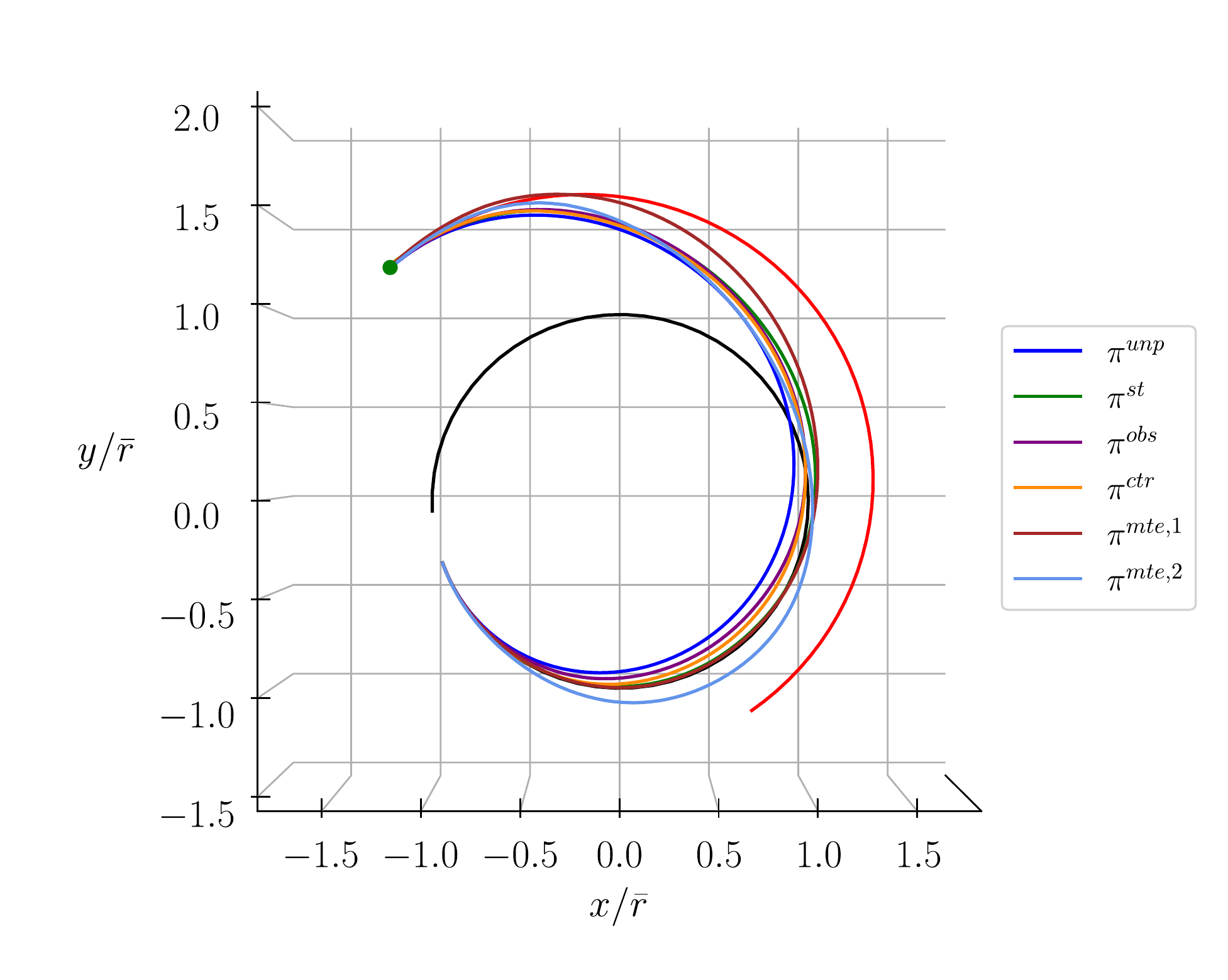}
    \caption{Earth-Mars trajectories corresponding to different robust policies. Differences with respect to the unperturbed policy trajectory are 
    up-scaled by  5 for illustration purposes.}
    \label{fig:robust_comp} 
\end{figure}

%
%
Figure~\ref{fig:robust_comp} shows the robust trajectories obtained after a training that lasts up to  $\SI{200}{\mega steps}$, that roughly corresponds to 10 $\div$ 12 computing hours.
For each case, only the best found solution during training (also accounting for the closed-loop behavior described in the next section) is reported.
To add robustness, these trajectories tend to approach Mars orbit in advance with respect to the optimal solution, so as to maximize the probability of meeting the terminal constraints even in presence of late perturbations and/or control errors.

\input{Tables/tab-robust}
%
Table~\ref{tab:nom} summarizes the main features 
of these trajectories, that are, the final spacecraft mass $m_f$, constraint violations $\Delta r_f$ and 
$\Delta v_f$, and cumulative reward $J$,
as well as some environment-specific training settings.
The solutions corresponding to a policy trained in a stochastic environment with perturbations on either state ($\pi^{st}$), observations ($\pi^{obs}$), or control direction and magnitude ($\pi^{ctr}$) satisfy the terminal constraints within the given tolerance ($10^{-3})$. In those cases,  robustness is obtained by sacrificing less than $1 \div 2$\% of the payload mass.  
Instead, the solutions $\pi^{mte,1}$ and $\pi^{mte,2}$, trained in the MTE environments,
tend to slightly overcome Mars orbit, since they account, even in the unperturbed scenario, for the possible presence of MTEs, whose probability of occurrence during training has been exaggerated in this work for research purposes (at least one MTE must occur in any environment realization). Also, the final spacecraft mass obtained in these two cases is considerably worse than in the previous ones.
In all presented cases, the error on the final velocity is zero. This result should not surprise the reader. In fact, the last $\Delta V$ is computed algebraically as a difference between the final spacecraft velocity and Mars velocity. Thus, the velocity constraint is automatically satisfied whenever this (computed) $\Delta V$ has a magnitude lower than the maximum admissible by the Sims-Flanagan model (see Eq.~\ref{eq:DVmax-f}).

\begin{figure}[!htbp]
    \centering
    \begin{subfigure}[t]{.49\textwidth}
    \centering 
       \includegraphics[width = 1\textwidth]{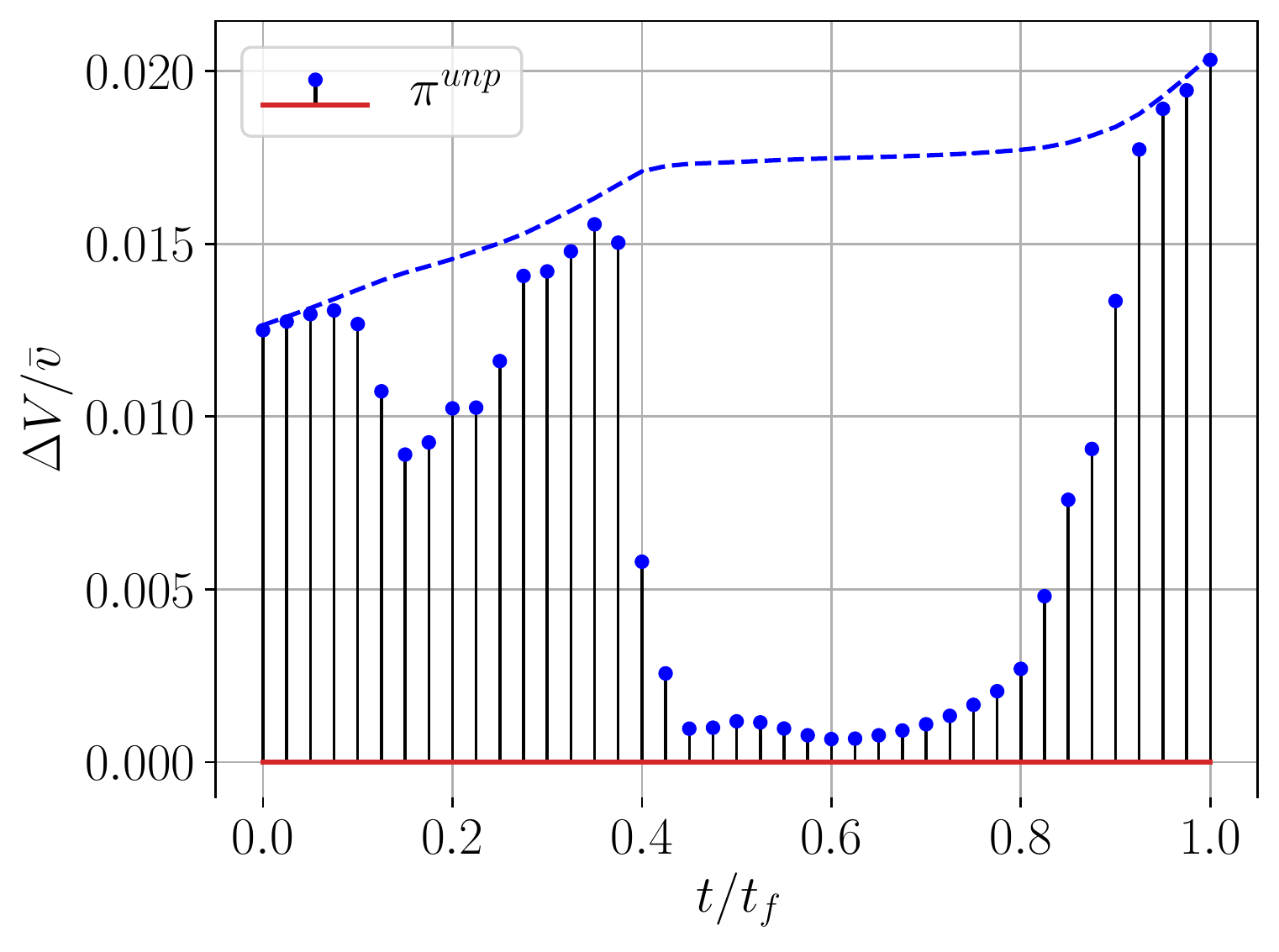} 
       \caption{Unperturbed policy.}
       \label{fig:dv_nom}
   \end{subfigure}%
    \hfill
    \begin{subfigure}[t]{.49\textwidth}
        \centering
        \includegraphics[width = 1\textwidth]{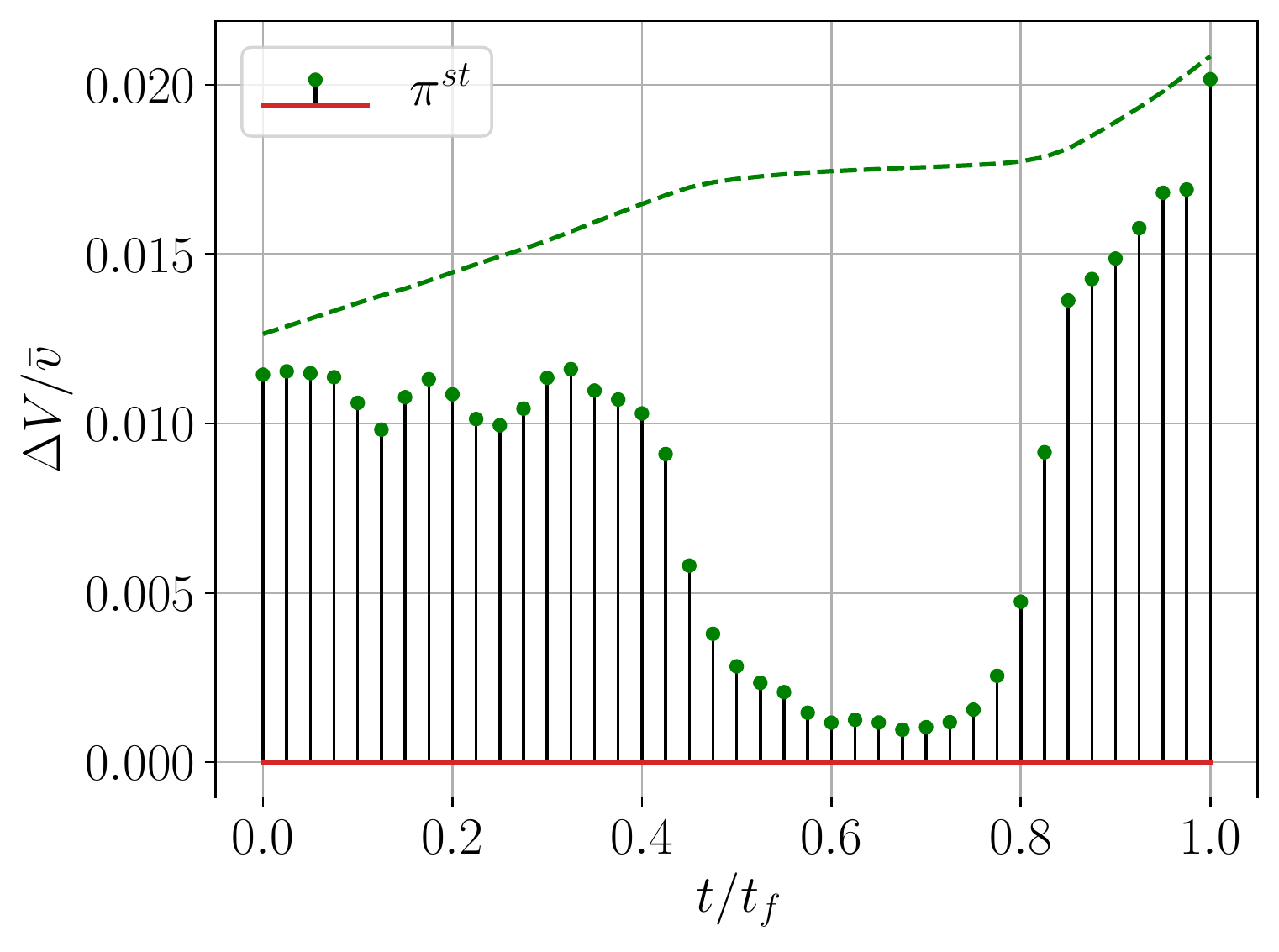}  
       \caption{State uncertainties.}
        \label{fig:dv_state}
    \end{subfigure}
     \begin{subfigure}[t]{.49\textwidth}
    \centering 
       \includegraphics[width = 1\textwidth]{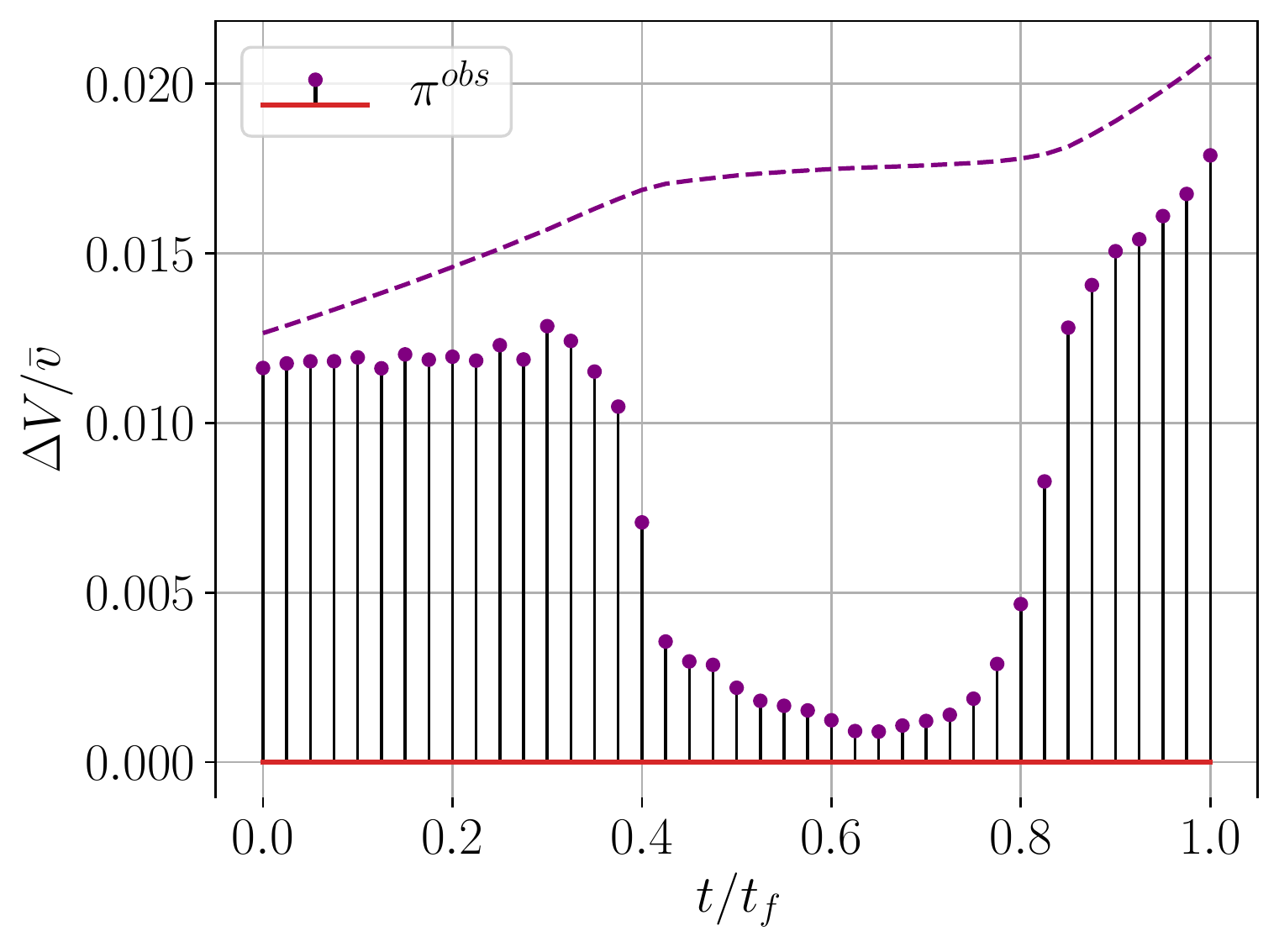} 
       \caption{Observation uncertainties}
       \label{fig:dv_obs}
   \end{subfigure}%
    \hfill
    \begin{subfigure}[t]{.49\textwidth}
        \centering
        \includegraphics[width = 1\textwidth]{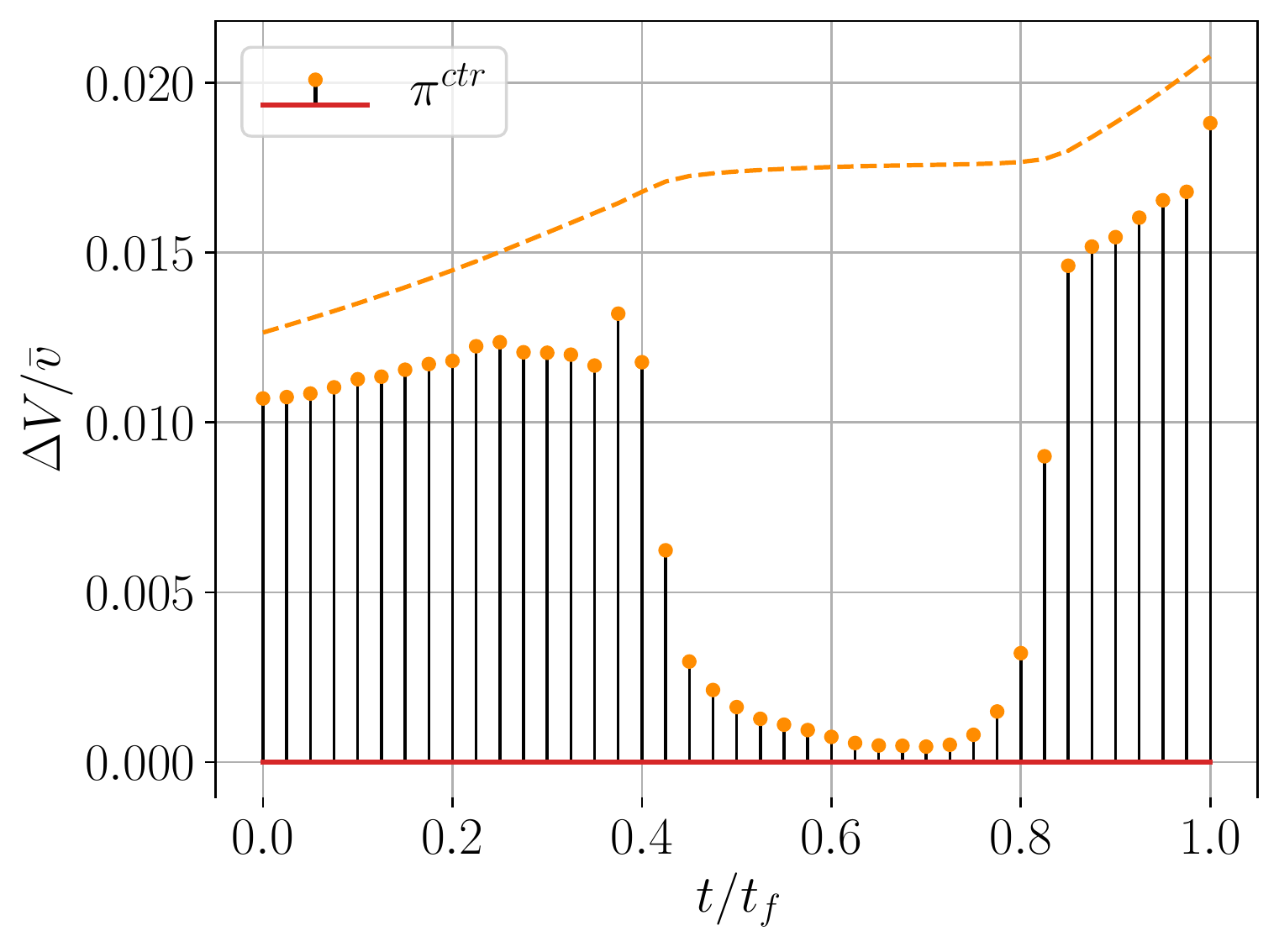}  
       \caption{Control uncertainties.}
        \label{fig:dv_con}
    \end{subfigure}
     \begin{subfigure}[t]{.49\textwidth}
    \centering 
       \includegraphics[width = 1\textwidth]{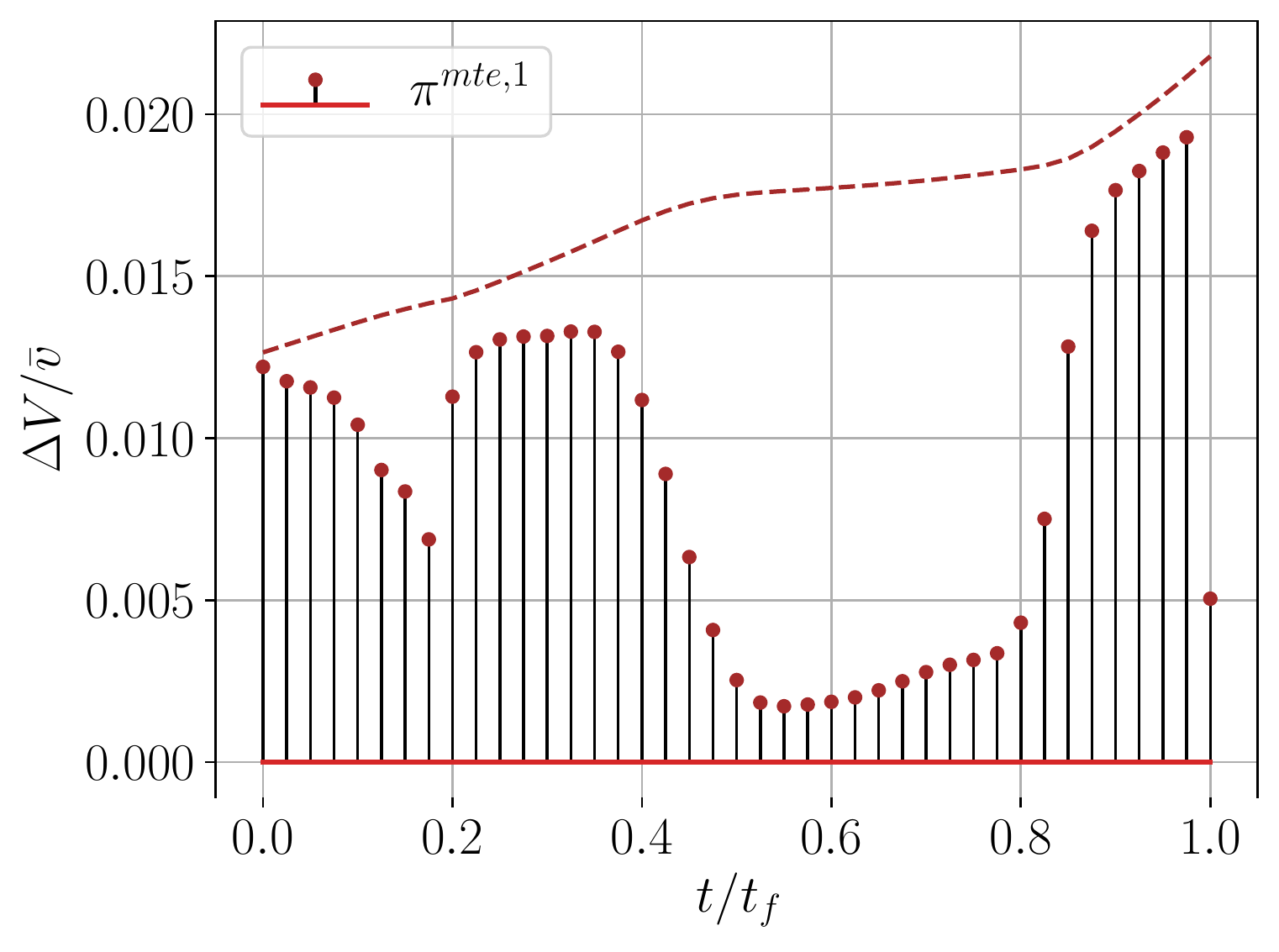} 
       \caption{Single MTE.}
       \label{fig:dv_mte}
   \end{subfigure}%
    \hfill
    \begin{subfigure}[t]{.49\textwidth}
        \centering
        \includegraphics[width = 1\textwidth]{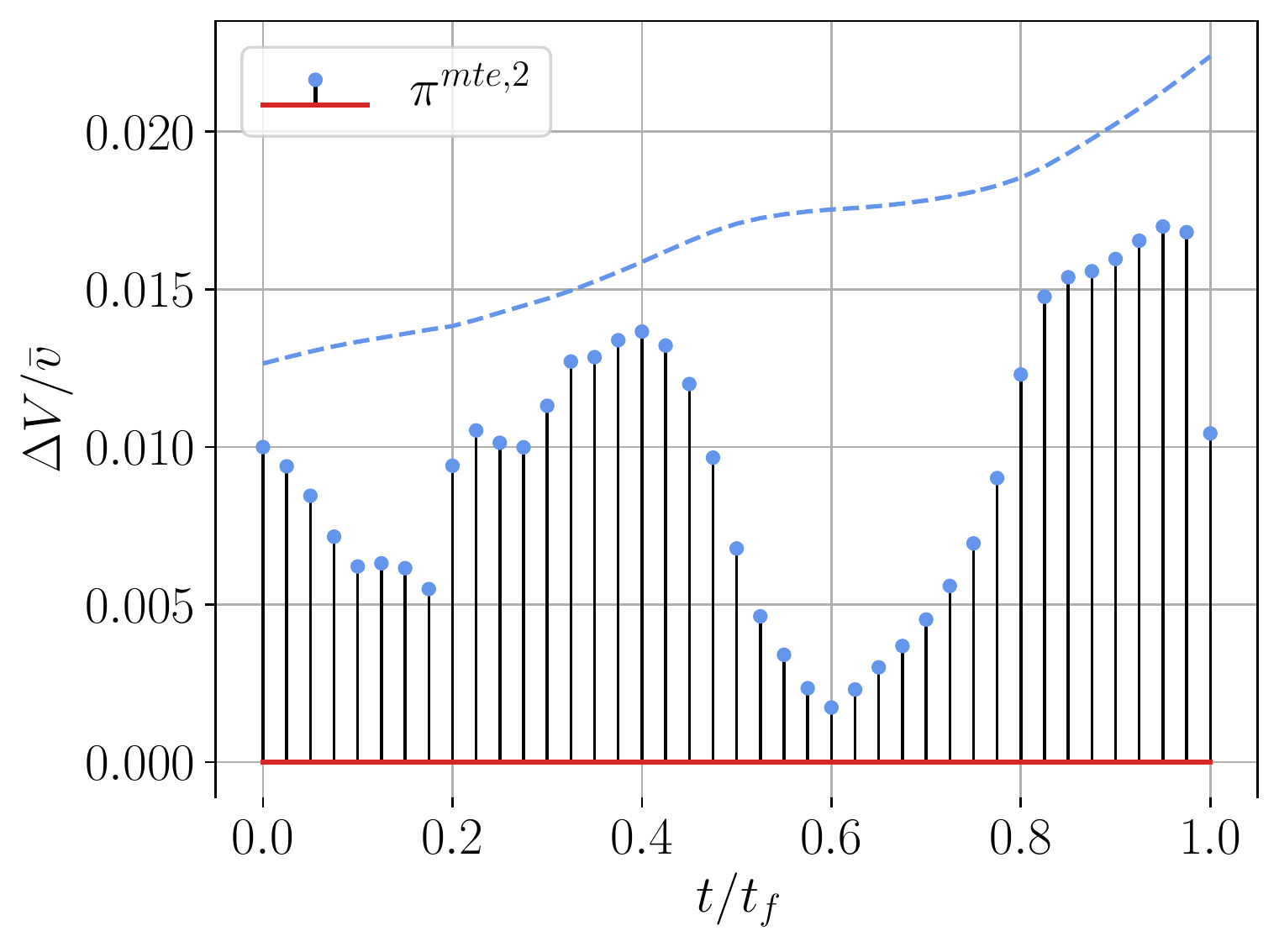}  
       \caption{Multiple MTEs.}
        \label{fig:dv_mtes}
    \end{subfigure}
    \caption{Magnitude of the $\pmb{\Delta V}$s along the robust trajectories. Dashed lines indicate the maximum $\pmb{\Delta V}$ admitted at each time step by Sims-Flanagan model.}
    \label{fig:DVrob}
\end{figure}

Figure~\ref{fig:DVrob} shows the distribution of the magnitude of the $\Delta V$s over the flight time for the different robust trajectories. 
Dashed lines indicate the maximum allowed $\Delta V$ at each step, according to the Sims-Flanagan model (see Eq.~\eqref{eq:DVmax-k}). 
As a general comment, the robust trajectories trained in the stochastic environments
show a lower $\Delta V$ magnitude at the beginning and at the end of the transfer, 
with respect to the optimal deterministic solution $\pi^{unp}$,
and, correspondingly, an higher magnitude in the central portion of the transfer.
This sub-optimal distribution of the thrust is responsible for the additional propellant consumption of the robust solutions.
%
%
Also, the applied $\Delta V$ is consistently lower than the maximum available in all cases except the unperturbed one, where an almost bang-off-bang pattern, which is the expected solution of this kind of optimal control problems, may be recognized.
This is a distinctive feature of robust trajectories, that must satisfy the constraint on the maximum admissible value of $\Delta V$, while leaving room for efficient correction maneuvers.
As a final remark, in the two solutions obtained with policies $\pi^{mte,1}$ and $\pi^{mte,2}$ the last $\Delta V$ is considerably smaller than in the other solutions, probably because 
the two policies try to ensure the compliance with the final velocity constraint regardless of the possible presence of a MTE near the final time.

\subsection{Closed-Loop Mission Analysis}
%
%
%
%
%
\begin{figure}[!htbp]
    \centering

    \hfill
    \begin{subfigure}[t]{.49\textwidth}
        \centering
        \includegraphics[width = 0.95\textwidth]{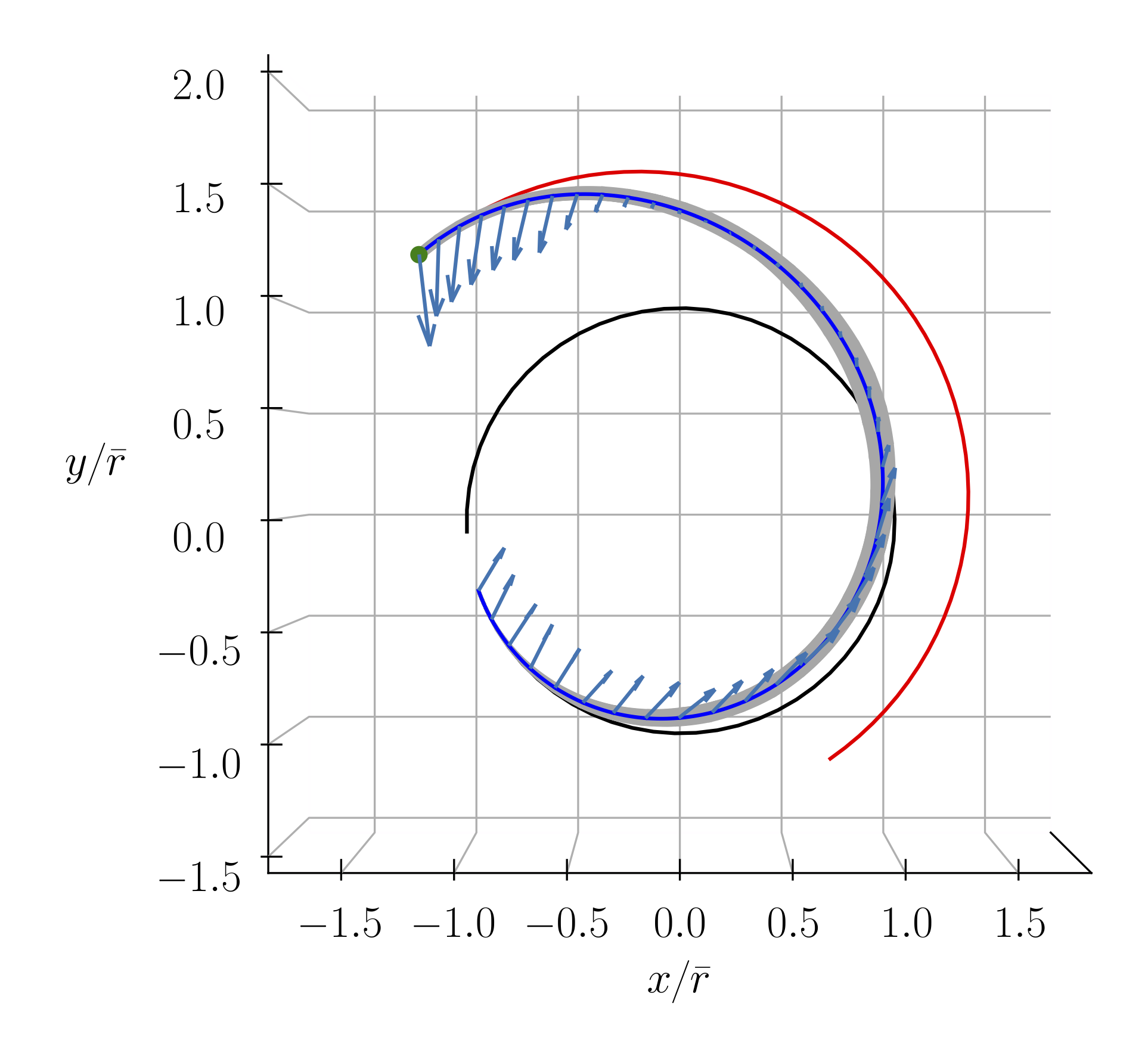}  
       \caption{$\pi^{st}$}
        \label{fig:MC_pert}
    \end{subfigure}
    \hfill
    \begin{subfigure}[t]{.49\textwidth}
    \centering 
       \includegraphics[width = 0.95\textwidth]{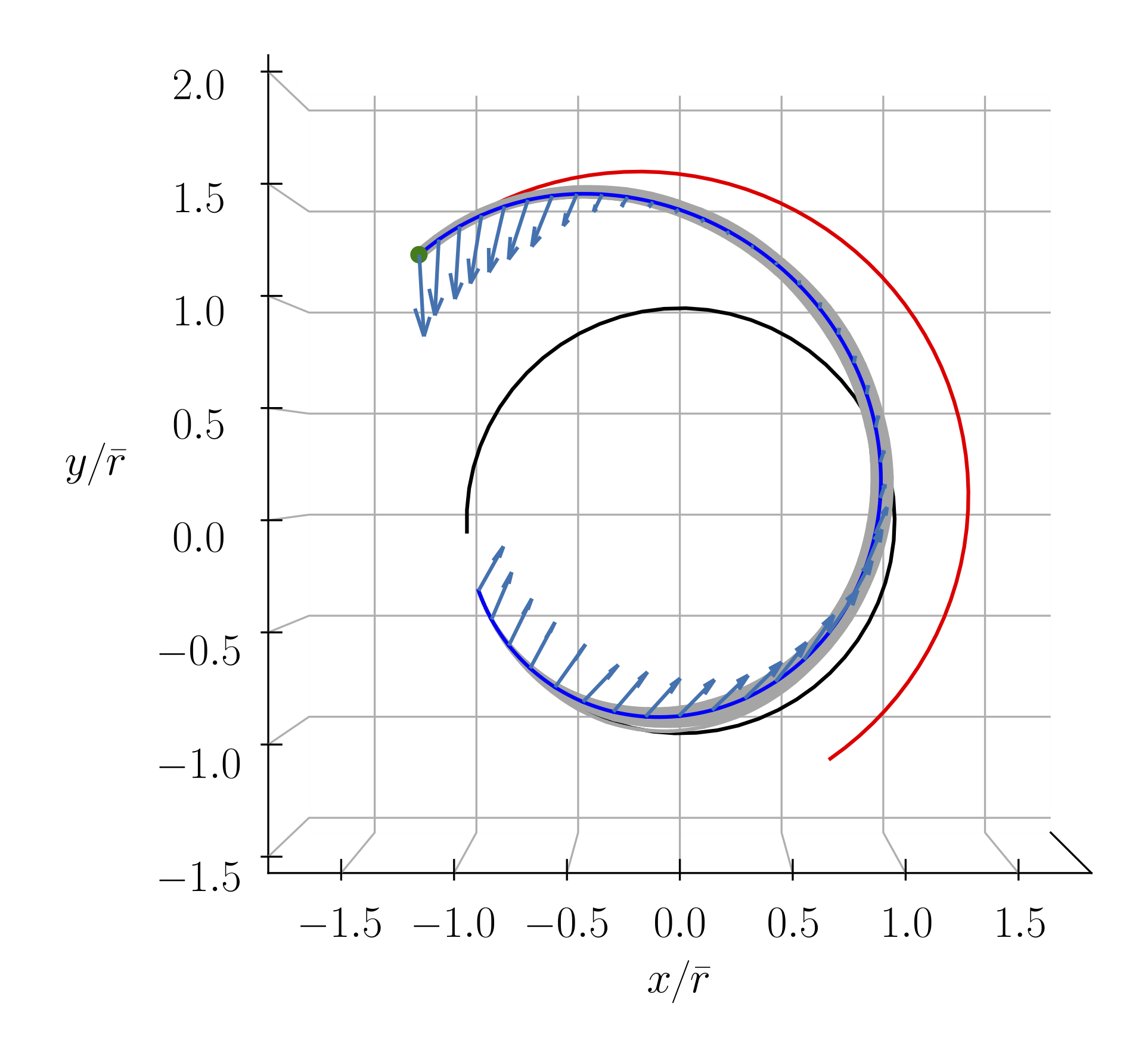} 
       \caption{$\pi^{obs}$}
       \label{fig:obs}
   \end{subfigure}%
    
    \begin{subfigure}[t]{.49\textwidth}
        \centering
        \includegraphics[width = 0.95\textwidth]{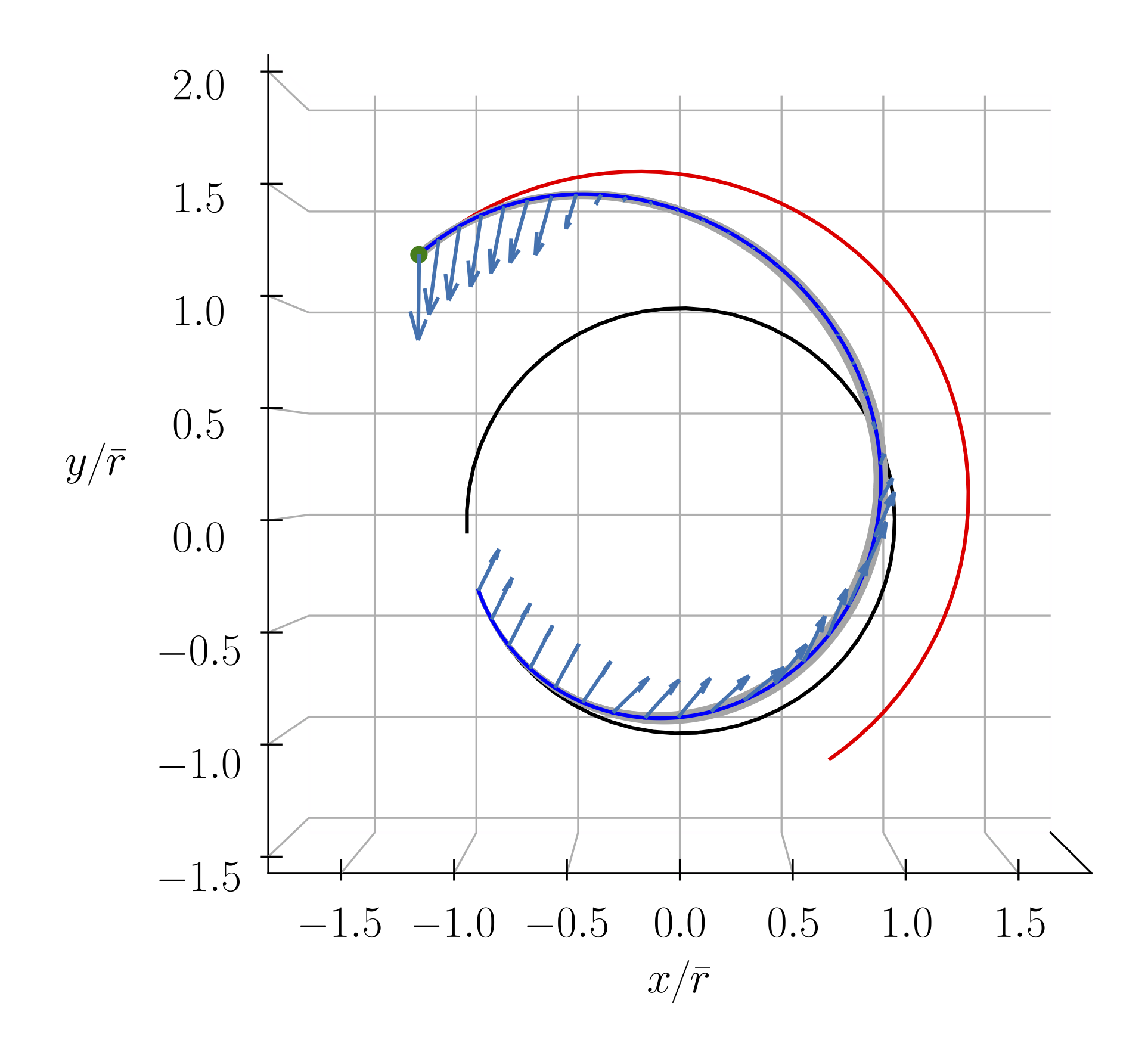}  
       \caption{$\pi^{ctr}$}
        \label{fig:control}
    \end{subfigure}
    \hfill
     \begin{subfigure}[t]{.49\textwidth}
    \centering 
       \includegraphics[width = 0.95\textwidth]{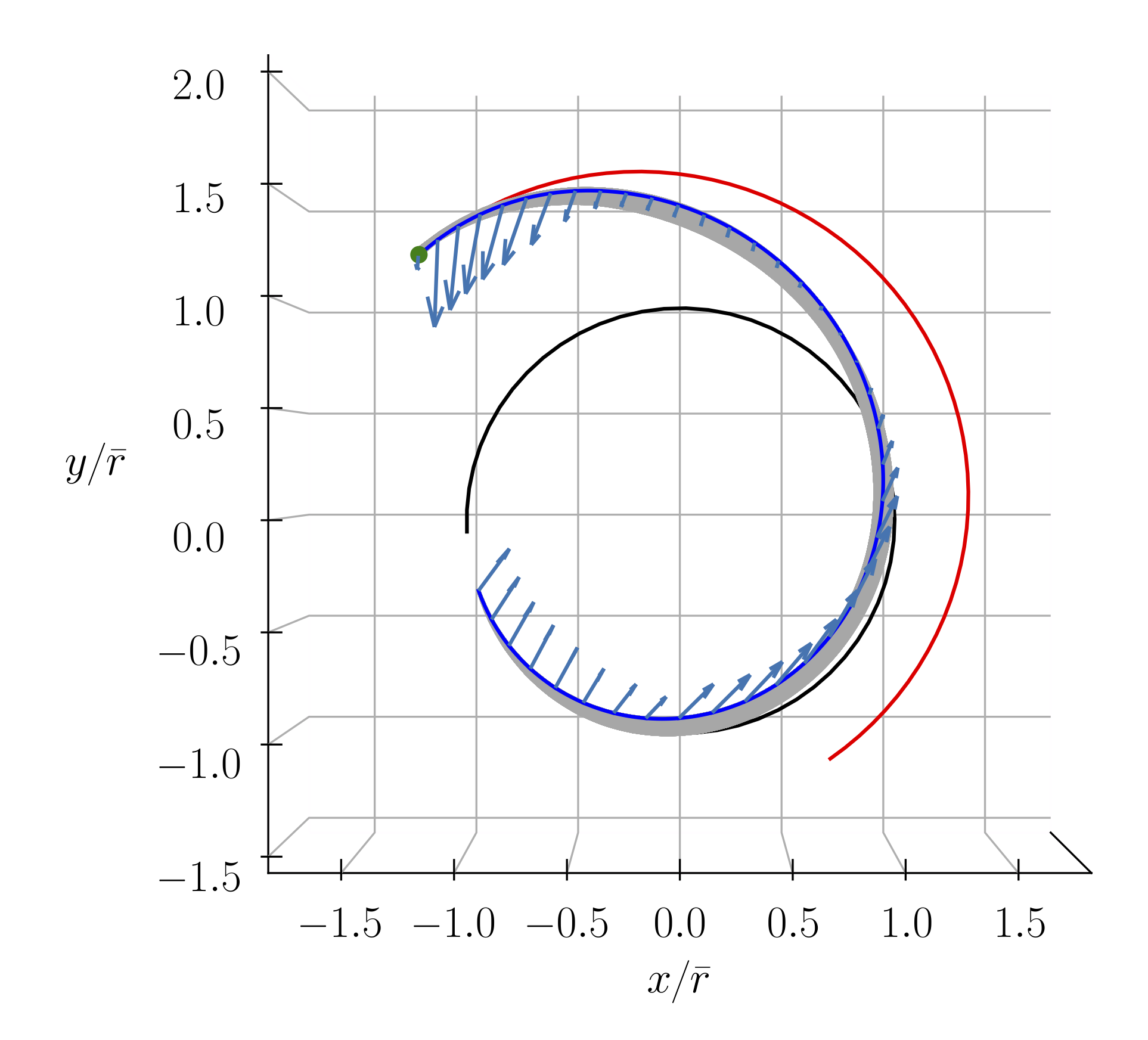} 
       \caption{$\pi^{mte,1}$}
       \label{fig:MTE}
   \end{subfigure}%
    
    \begin{subfigure}[t]{.49\textwidth}
        \centering
        \includegraphics[width = 0.95\textwidth]{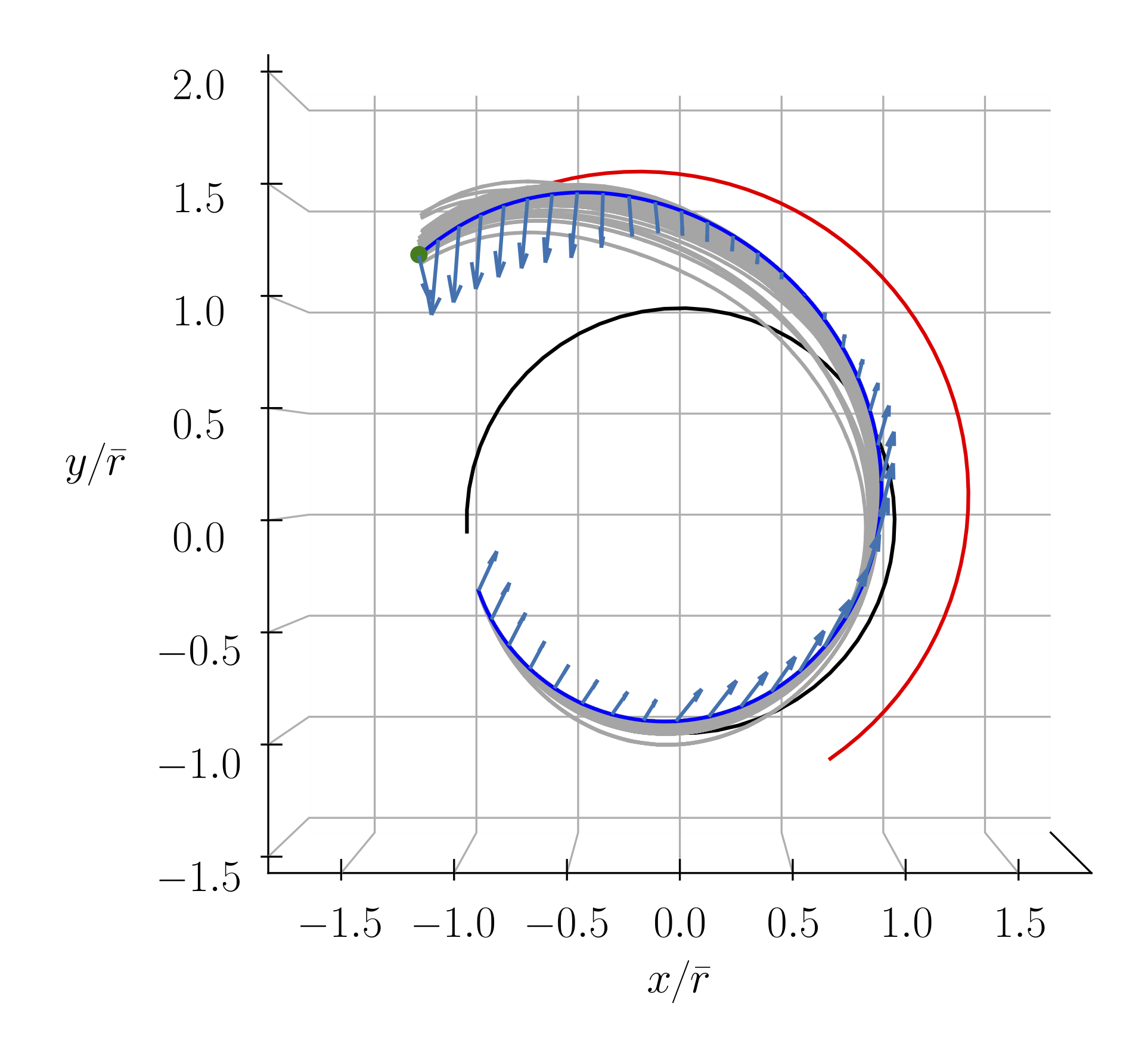}  
       \caption{$\pi^{mte,2}$}
        \label{fig:MTEs}
    \end{subfigure}
    \hfill
        \begin{subfigure}[t]{.49\textwidth}
    \centering 
       \includegraphics[width = 0.95\textwidth]{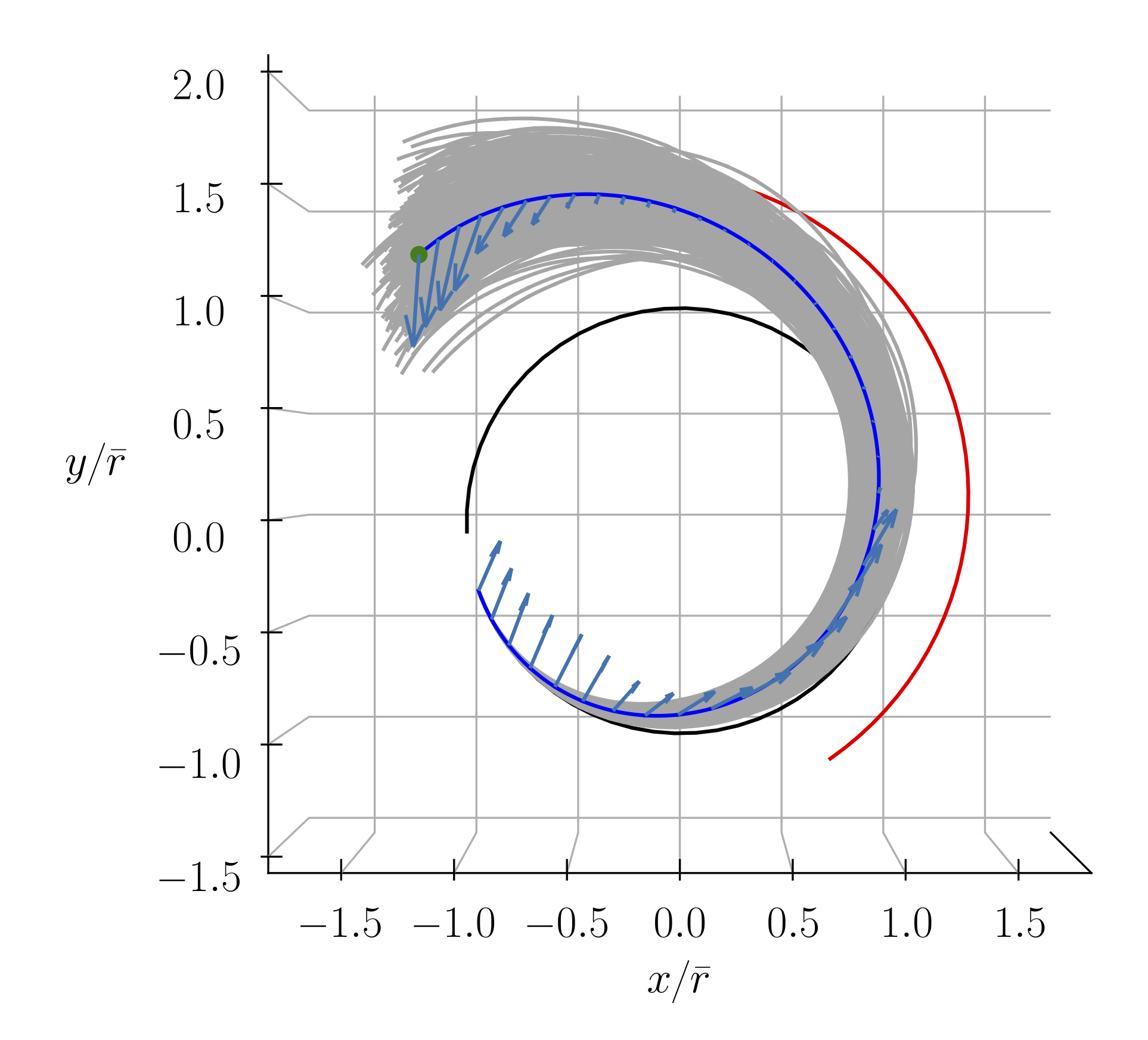} 
       \caption{$\pi^{unp}$}
       \label{fig:MC_det}
   \end{subfigure}%
    \caption{Monte Carlo simulations realized by running each policy in the correspondingly-perturbed environment, except for the unperturbed policy, which is run in the state-perturbed one. Differences are exaggerated by a factor 5 for illustration purposes.}
    \label{fig:MCs}
\end{figure}

The closed-loop performance of the \gncnet{s} in their respective (stochastic) training environment are measured by using a Monte Carlo approach that consists in running each state-feedback policy in 500 randomly-generated environment realizations, or episodes. 
Figure~\ref{fig:MCs} shows the results of the Monte Carlo campaigns in terms of spacecraft trajectory in each randomly-generated episode. Specifically, in each figure, the dark-blue line represents the robust trajectory, with the light-blue arrows indicating the $\Delta V$s, while each gray line represents the trajectory obtained in one of the randomly-generated episodes. 
One may notice that, for policies $\pi^{st}$, $\pi^{obs}$, $\pi^{ctr}$ and $\pi^{mte,1}$, the Monte Carlo generated trajectories have a greater dispersion in the central part of the mission, which tends to reduce, and disappear almost entirely, while approaching the final time, because of the imposed terminal constraints. This is not completely true for the case of multiple MTEs, where a small number of trajectories (15 out of 500) clearly miss the target by far.

\input{Tables/tab-MC}
Table~\ref{tab:MC} shows that RL is able to cope with all these different stochastic scenarios 
effectively. 
Indeed, despite the severity of the considered perturbations/uncertainties, the success rate (that is, the percentage of solutions that meet final constraints within the prescribed accuracy) is rather high, over 70\% in most cases and up to 80\% when only additive Gaussian state perturbations are considered.
%
%
For the sake of comparison, we also reported the results obtained by running policy $\pi^{unp}$ in the state-perturbed stochastic environment (Fig.~\ref{fig:MC_det}).
While the differences between the robust trajectories corresponding to policy $\pi^{unp}$ and $\pi^{st}$ seem minimal, the effects in the closed-loop simulations are apparent.
Indeed, in none of the episodes the policy $\pi^{unp}$ was able to reach the imposed accuracy on terminal state, while policy $\pi^{st}$ succeeds in the $80\%$ of the cases.
Similar results are obtained when running the policy $\pi^{unp}$ in any of the proposed perturbed (stochastic) environments.
More precisely, the success rate is zero for both the state- and observation-uncertainty environments, while in case of control uncertainties on thrust magnitude/direction is 8\%, and almost double in case of single (18.8\%) or multiple (16.8\%) MTEs.

\begin{figure}[!htbp]
    \centering
    \begin{subfigure}[t]{.47\textwidth} 
    \centering 
       \includegraphics[width = 1\textwidth]{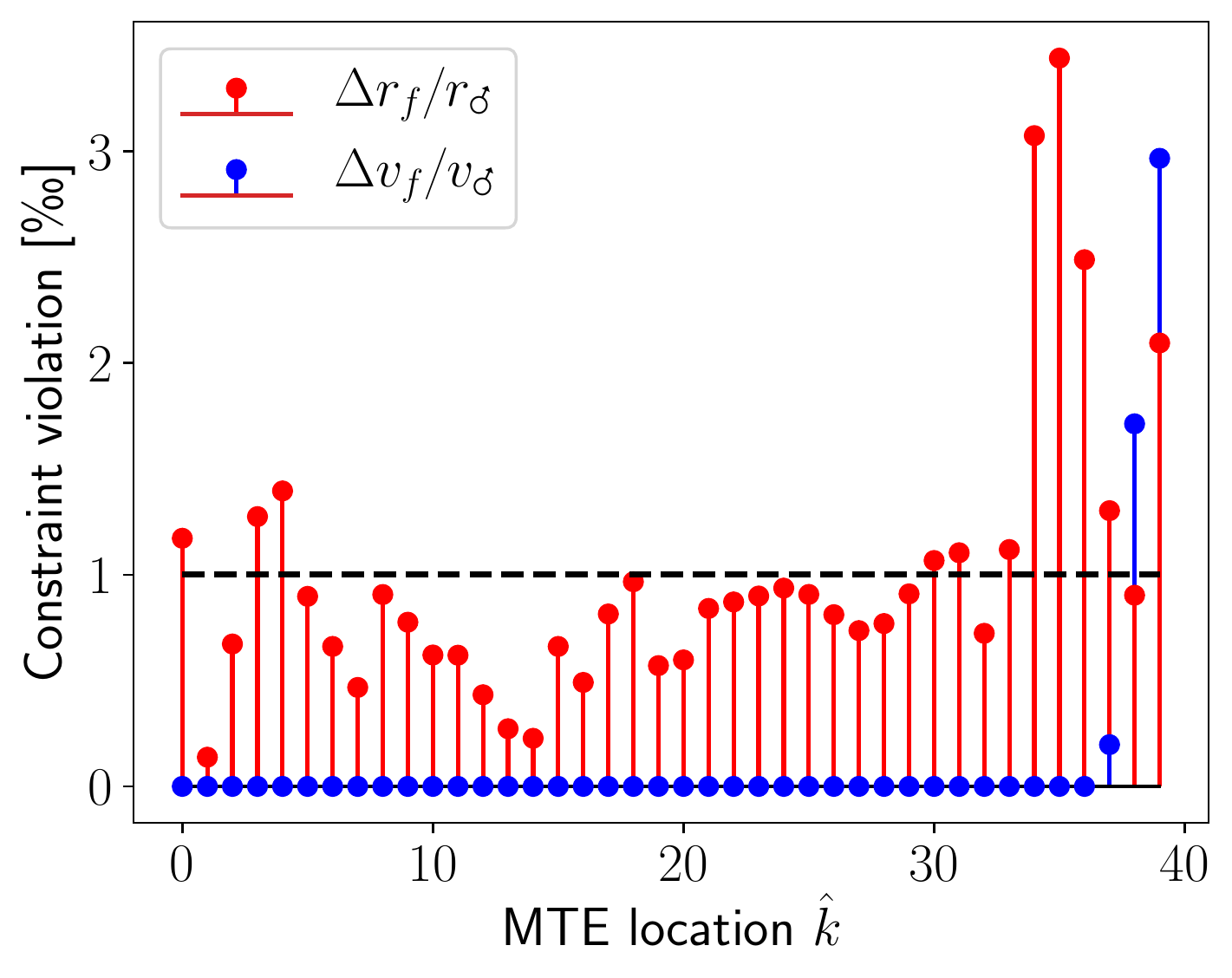} 
       \caption{Terminal constraint violation.}
       \label{fig:errors}
   \end{subfigure}%
    \hfill
    \begin{subfigure}[t]{.51\textwidth} 
        \centering
        \includegraphics[width = 1\textwidth]{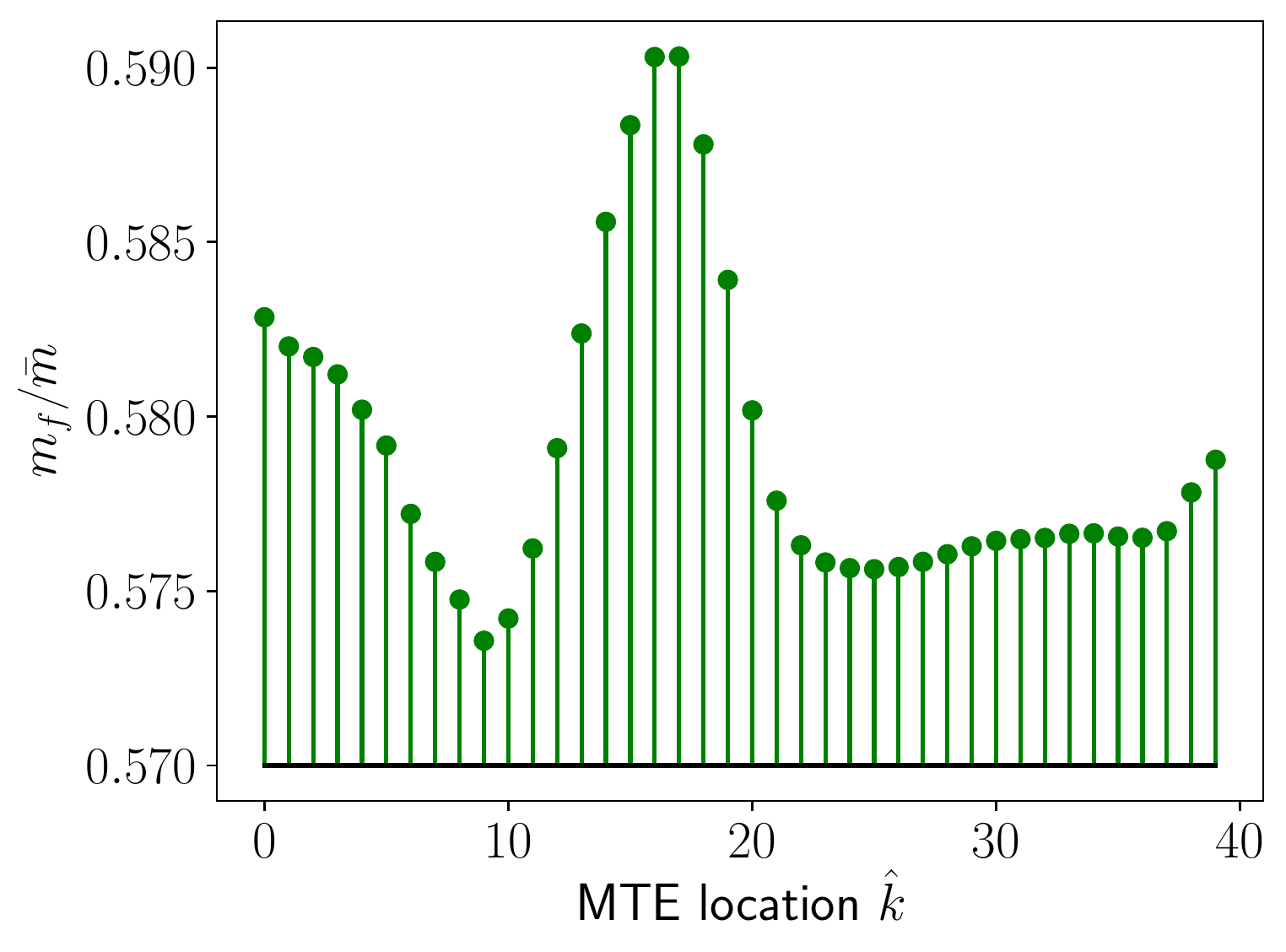}  
       \caption{Final spacecraft mass.}
        \label{fig:mass}
    \end{subfigure}
    \caption{Terminal constraint violation (a) and final spacecraft mass (b) obtained with policy $\pmb{\pi^{mte,1}}$ by varying the MTE location $\pmb{\hat{k}}$.}
    \label{fig:mte_analysis}
\end{figure}

The preliminary results found with policies $\pi^{mte,1}$ and $\pi^{mte,2}$ 
stressed that RL performance deteriorates substantially in the presence of MTEs.
By looking at Figure~\ref{fig:errors}, it is clear that, in most cases ($69.8\%$ with a single MTE, $70.4\%$ with multiple MTEs) the policy manages to recover from the complete absence of thrust, and meets the final constraints within the imposed tolerance ($10^{-3}$). However, in a few unfortunate scenarios, the MTE occurs in ``crucial points'' of the trajectory, that is, near final time, and the policy is not able to compensate for the missing $\Delta V$s in any way. As a result, the terminal constraints cannot be met. This fact is confirmed by the high variance obtained on the constraint violation in both mission scenarios (single and multiple MTEs). Analogously, the drop in payload mass (Fig.~\ref{fig:mass}) highlights what are the most critical points in terms of thrust efficiency. The looser satisfaction of the terminal constraints is deemed responsible for the final rise in the achieved payload mass and might be misleading.

%% file: Tables/tab-robust.tex
\begin{table}[!htbp]
    \caption{Robust trajectory overview.    
    }
   \label{tab:nom}
        \centering 
   \begin{tabular}{l c c c | c c c c} 
   \hline 
     \multirow{2}{*}{Policy} &\multicolumn{3}{c|}{Settings} & \multicolumn{4}{c}{Results} \\
      & $T$ & $n_{env}$ & $n_b$ & $m_f,\,\,\si{\kilo\gram}$ & $\Delta r_f/r_\Mars,\,\,$\textperthousand  & $\Delta v_f/v_\Mars,\,\,$\textperthousand & $J$  \\
      \hline 
      $\pi^{unp}$& $48M$ & $8$ & $4$ & $600.23$ & $0.999$ & $0$ & $-0.3998$ \\
      $\pi^{st}$ & $112M$ & $8$ & $4$ & $588.41$ & $0.767$ & $0$ & $-0.4116$ \\
      $\pi^{obs}$& $200M$ & $8$ & $4$ & $591.48$ & $0.782$ & $0$ & $-0.4085$ \\
      $\pi^{ctr}$& $128M$ & $16$ & $4$ & $591.40$ & $0.814$ & $0$ & $-0.4086$ \\
      $\pi^{mte,1}$& $80M$ & $16$ & $4$ & $575.98$ & $3.689$ & $0$ & $-0.5585$ \\
      $\pi^{mte,2}$& $128M$ & $16$ & $8$ & $555.87$ & $2.273$ & $0$ & $-0.5078$ \\
      \hline 
   \end{tabular}
\end{table}

%% file: Tables/tab-MC.tex
\begin{table}[!h]
    \caption{Results of the Monte Carlo simulations in terms of mean value (mean), standard deviation (std), and success rate (SR).}
\begin{adjustwidth}{-1in}{-1in}
   \label{tab:MC}
        \centering 
   \begin{tabular}{l  c c   c c   c c  c } 
      \hline 
      \multirow{2}{*}{Policy} & \multicolumn{2}{c }{$m_f,\,\,\si{\kilo\gram}$} & \multicolumn{2}{c }{$\Delta r_f/r_\Mars,\,\,$\textperthousand}  & \multicolumn{2}{c }{$\Delta v_f/v_\Mars,\,\,$\textperthousand} & \multirow{2}{*}{SR, \%}  \\
       & mean  & std &  mean & std &  mean & std &  \\
      \hline 
      $\pi^{st}$ & $578.04$ & $9.48$ & $0.742$ & $0.322$ & $0.0403$ & $0.211$ & $80.0$ \\
      $\pi^{obs}$ & $580.94$ & $9.53$ & $0.807$ & $0.343$ & $0.0502$ & $0.299$ & $70.6$ \\
     $\pi^{ctr}$ & $591.13$ & $0.685$ & $0.846$ & $0.426$ & $0.0620$ & $0.270$ & $69.0$ \\
     $\pi^{mte,1}$ & $579.02$ & $4.44$ & $0.980$ & $0.667$ & $0.0987$ & $0.438$ & $69.8$ \\
      $\pi^{mte,2}$ & $557.32$ & $4.77$ & $1.266$ & $2.048$ & $0.645$ & $3.451$ & $70.4$ \\
      \hline
   \end{tabular}
  \end{adjustwidth}
\end{table}